\documentclass[final]{cvpr}

\usepackage{times}
\usepackage{epsfig}
\usepackage{graphicx}
\usepackage{amsmath}
\usepackage{amssymb}
\usepackage{booktabs}
\usepackage{multirow}
\usepackage[ruled,vlined]{algorithm2e}
\usepackage{algorithmic}
\usepackage{listings}
\usepackage{color}
\usepackage{subfiles}

\usepackage{comment}

\newcommand{\B}{\fontseries{b}\selectfont}
\definecolor{brown}{RGB}{128, 76, 37}
\definecolor{dc}{RGB}{255, 122, 211}
\usepackage[pagebackref=true,breaklinks=true,colorlinks,bookmarks=false]{hyperref}

\def\cvprPaperID{787} 
\def\confYear{CVPR 2021}

\begin{document}

\title{
Bridging In- and Out-of-distribution Samples for Their Better Discriminability
}

\author{
Engkarat Techapanurak$^{1}$~~~~~ Anh-Chuong Dang$^{1}$ ~~~~~Takayuki Okatani$^{1,2}$\\
$^1$Graduate School of Information Sciences, Tohoku University ~~~~ $^2$RIKEN Center for AIP\\
Sendai, 980-8579, Japan \\
{\tt\small \{engkarat, anhcda, okatani\}@vision.is.tohoku.ac.jp}
}

\maketitle

\begin{abstract}
This paper proposes a method for OOD detection. Questioning the premise of previous studies that ID and OOD samples are separated distinctly, we consider samples lying in the intermediate of the two and use them for training a network. We generate such samples using multiple image transformations that corrupt inputs in various ways and with different severity levels. We estimate where the generated samples by a single image transformation lie between ID and OOD using a network trained on clean ID samples. To be specific, we make the network classify the generated samples and calculate their mean classification accuracy, using which we create a soft target label for them. We train the same network from scratch using the original ID samples and the generated samples with the  soft labels created for them. We detect OOD samples by thresholding the entropy of the predicted softmax probability. The experimental results show that our method outperforms the previous state-of-the-art in the standard benchmark tests. We also analyze the effect of the number and particular combinations of image corrupting transformations on the performance. 
\end{abstract}

\section{Introduction} \label{sec::introduction}

Detecting out-of-distribution (OOD) samples, i.e., samples from a distribution other than the distribution of the samples used for training (called `in-distribution (ID)' samples), is a vital problem to cope with when deploying neural networks in real-world applications. There are many studies on the problem so far. The difficulty with the OOD detection lies in the requirement to distinguish ID and OOD samples by learning only ID samples. A natural approach is to formulate the problem as anomaly detection. Several methods \cite{lee2018simple,yu2020compression,sastry2020gram_matrices,zisselman2020resflow} utilize the intermediate layer activation of a network that is trained on ID samples;
they model the distribution of ID samples in the feature space and detect OOD samples as anomalies. They perform fairly well in some cases but show limitations \cite{shafaei2018biased,techapanurak2019hyper,hsu2020generalize}.

To gain further performance, it seems necessary to have a better feature space, in which OOD samples are more clearly distinguished from ID samples. Evidence for the necessity is that using a network pre-trained on a large dataset (e.g., ImageNet) makes OOD detection easier \cite{hendrycks2019pretrain}. Furthermore, a method using cosine similarity to model class probabilities, as with metric learning methods, has been proposed \cite{techapanurak2019hyper,hsu2020generalize}, showing promising results. There are several methods \cite{liang2017enhancing,Yu2019UnsupervisedOD,sastry2020gram_matrices,yu2020compression} that map the layer activation from a trained network into a good feature to enable more accurate OOD detection.

Outlier Exposure (OE) \cite{hendrycks2018outlier_oe} is yet another approach to OOD detection that can be thought of as aiming at the same goal. It uses an available OOD dataset at training time, which hypothetically does not need to match the OOD sample distribution we will encounter in practice. A network is trained to classify an input to its true class if it is ID and `none of the classes,' (i.e., identical probabilities for all the ID classes) if it is OOD. 
It aims to learn a better internal representation that helps identify unseen OOD samples accurately. While it shows good experimental performance, its success inevitably depends on the (dis)similarity between the assumed and the real OOD samples, which is hard to quantify in experiments; thus, its real-world performance remains unclear.

In this paper, we question the premise of previous studies that ID and OOD samples are separated distinctly, proposing a new OOD detection method. As mentioned above, OE regards any sample as either an ID or an OOD sample in an exclusive manner. In contrast, we consider samples in the intermediate between the two, i.e., those having an OOD likelihood between 0 to 100\%. We consider these samples to have soft labels and train a network using them.

The problem is how to get such intermediate samples as well as their soft labels. To do this, we apply synthetic image corruption to ID samples, creating new samples lying between ID and OOD. The underlying thought is that applying very severe image corruption to ID samples will make them turn to OOD, as their semantic contents will be lost. 
On the other hand, less severe corruption will create samples maintaining their contents; their ID/OOD likelihoods will be in the range of 0 to 1. 

To provide soft labels for these intermediate samples, we use a network trained on the ID samples alone in the standard fashion. Specifically, we apply an image corrupting transformation to all the samples of the ID training set and then input the transformed samples to the above network, calculating their mean classification accuracy. We then use it to create a soft label. Concretely, we design the soft label for a sample to have the mean classification accuracy as the probability of its true class and a constant probability for all other classes. This method creates a single soft label for a single corrupting transformation. Ideally, we want to create soft labels distributing uniformly in the range between ID and OOD. For this purpose, we employ image corrupting methods that were developed to create the ImageNet-C dataset \cite{hendrycks2019robustness}, which consists of fifteen different types of image corruption, each with five severity levels, i.e., 75 corruption methods in total. Leveraging their diversity, we create multiple soft labels sampling as densely as possible in the intermediate region between ID and OOD. 

\begin{figure}[t]
    \centering
    \includegraphics[width=0.9\linewidth]{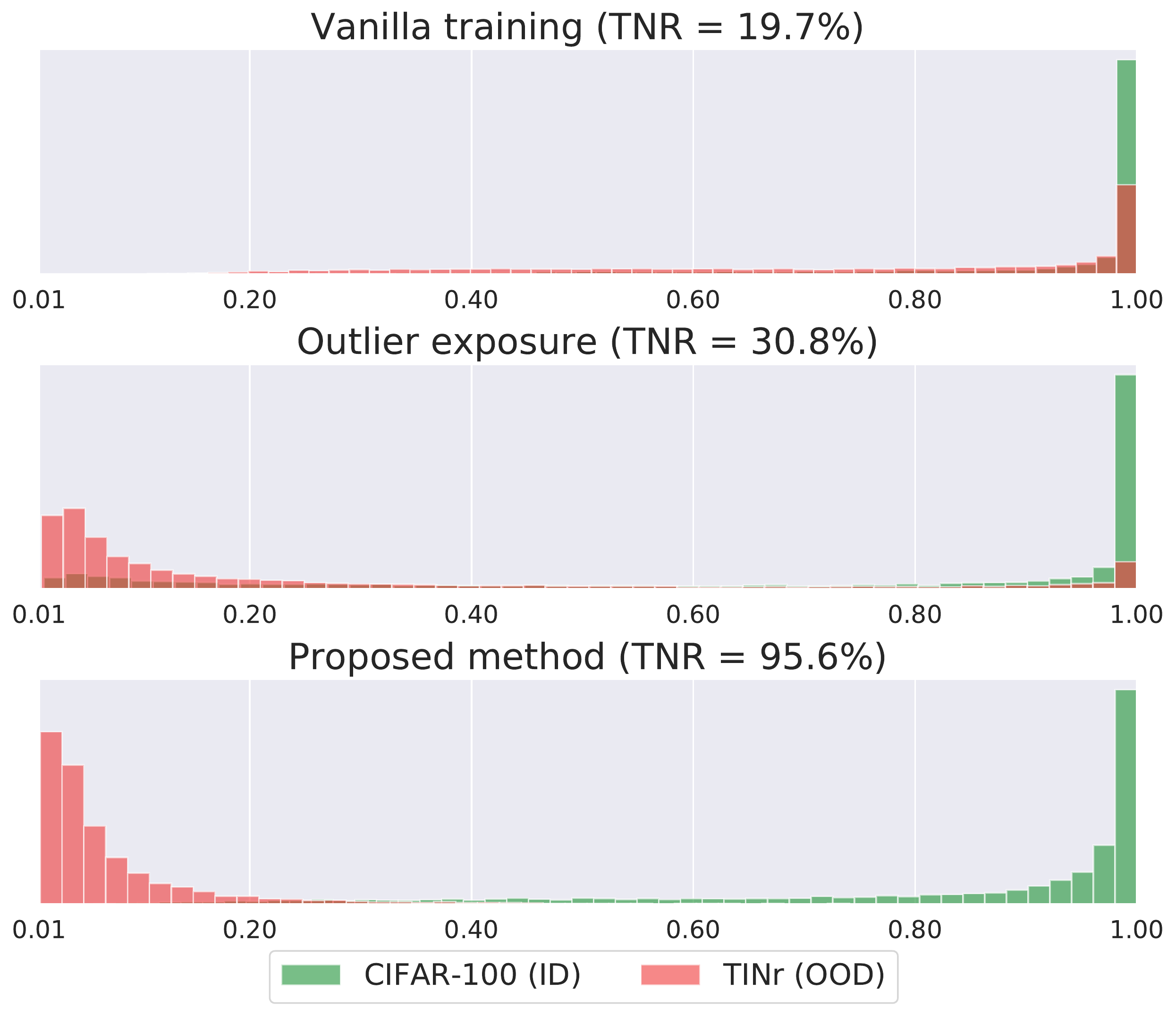}
    \caption{Histogram of the predicted confidence (i.e., maximum softmax probability) by a network (i.e., Wide Resnet 40-4) for in-distribution (ID) and out-of-distribution (OOD) samples. {\B First row:} The network with the vanilla training. {\B Second row:} Outlier exposure \cite{hendrycks2018outlier_oe}. {\B Third row:} Proposed method. ID and OOD are  CIFAR-100 and the resized Tiny ImageNet (TINr). {\em TNR at TPR 95\%} are shown in the parentheses.}
    \label{fig::hist_id_ood}
\end{figure}

Training a standard CNN using the generated training samples with soft labels along with the original ID samples makes the CNN learn an improved internal representation, which separates ID samples and unseen OOD samples more clearly. Figure \ref{fig::hist_id_ood} shows the distributions of ID samples and OOD samples in the space of the predicted confidence of the same network trained differently. As a result, our method achieves state-of-the-art performance in the standard benchmark tests of OOD detection. 

\section{Related Work} \label{sec::related_work}

\subsection{OOD Detection}

The maximum softmax probability (MSP), also called confidence, can be thought of as an estimate of the prediction's uncertainty. Using it with simple thresholding is a strong baseline for OOD detection \cite{hendrycks2016baseline}, and many studies have proposed its extensions to improve detection accuracy. \cite{liang2017enhancing,devries2018learning,vyas2018ensemble_leave_out,hendrycks2018outlier_oe,Yu2019UnsupervisedOD}.

It is shown that the addition of a small perturbation to input that maximizes the confidence improves OOD detection accuracy  \cite{liang2017enhancing,hsu2020generalize}. A method adding a separated branch to the network learning to predict the confidence of prediction is proposed \cite{devries2018learning}. It is also proposed to use the cosine similarity to compute logits instead of the ordinary linear transformation before softmax function \cite{techapanurak2019hyper,hsu2020generalize}, yielding high performance. Other studies consider an ensemble of networks \cite{lakshminarayanan2017simple,vyas2018ensemble_leave_out}, or considers a different problem setting 
\cite{Yu2019UnsupervisedOD}.

Another group of methods formulates OOD detection as anomaly detection. They model the distribution of normal data (i.e., ID samples) in the space of some feature, which is intermediate layer activation \cite{lee2018simple} or its transformation by some mapping \cite{yu2020compression,sastry2020gram_matrices,zisselman2020resflow}. They then detect outliers of the distribution as OOD. 

Some of the above methods (e.g.,  \cite{liang2017enhancing,lee2018simple,zisselman2020resflow}) assume the accessibility to the true OOD samples if only a few. These methods have a few hyperparameters, which often significantly impact the final performance; thus, they determine them using the available OOD samples. Recent studies question this approach, as the true OOD samples are usually not accessible in practice \cite{shafaei2018biased,Yu2019UnsupervisedOD,techapanurak2019hyper,hsu2020generalize,yu2020compression,sastry2020gram_matrices,ruff2020unifying_anomaly}.

There are also studies treating OOD detection more like anomaly detection; they do not use the model that has learned the ID class classification task. These studies model the distribution of ID samples using generative models, such as GAN \cite{lee2017training}, PixelCNN \cite{ren2019likelihood}, and Normalizing Flow \cite{nalisnick2018glowOOD}. In \cite{Choi2020Novelty}, the authors synthetically blur input images and use the Random Network Distillation \cite{burda2018exploration} to detect the OOD.
However, these methods generally show inferior performance than the above methods that utilize the representation learned through the ID task training.

Yet another approach to detect OOD samples is to use the uncertainty of the prediction. We can think of the confidence-based methods mentioned above as following this approach. Methods with a more solid theoretical foundation are those based on Bayesian neural networks \cite{gal2017dropout,mcclure2016robustly,gast2018lightweight,azizpour2018batch_norm_bayesian}. However, these approaches do not show competitive performance to the above methods. 

\subsection{Soft Labels}

In the standard setting of multi-class classification, target labels are represented as hard labels and used for training. While this is reasonable considering the nature of classification tasks, researchers have employed soft target labels for several purposes. One is the label smoothing. Since it was introduced to train Inception-v2 \cite{szegedy2016rethinking,he2019bag}, many studies have employed this trick, aiming at performance improvement; \cite{pereyra2017regularizing_label_smoothing,zoph2018learning_transferable}, to name a few. Recently, M\"uller et al. \cite{muller2019label_smoothing_help} showed detailed analyses of the (in)effectiveness of label smoothing. Another use of soft target labels is seen in Knowledge Distillation \cite{hinton2015distilling}. A student network learns the soft labels provided by its teacher. It is shown in \cite{muller2019label_smoothing_help} that training the teacher with label smoothing worsens the student's performance. Soft labels are also used in the methods for dealing with label noise. Several methods estimate the confusion matrix defined between the network prediction and the provided noisy labels \cite{ren2018learning}. Others estimate the true labels of training samples during the training of a network \cite{tanaka2018joint,yi2019probabilistic}. We may think these methods train networks using soft target labels. Some methods for data augmentation also use soft target labels. An example is Mixup \cite{zhang2018mixup}, which interpolate two training samples by computing the weighted sum of not just inputs but their labels, yielding soft labels.

\section{Proposed Method} \label{sec::proposed_method}

\subsection{Revisiting Outlier Exposure} \label{sec::revisiting_outlier_exposure}

Outlier Exposure (OE) is a method for OOD detection  proposed by Hendrycks et al.~\cite{hendrycks2018outlier_oe}. It uses available OOD datasets for the training of a model to increase its sensitivity to unseen OOD samples. It is a general framework and they consider multi-class classification and density estimation. We consider the former here. 

The method considers three distributions of samples, $\mathcal{D}_{\mathrm{in}}$, $\mathcal{D}_{\mathrm{out}}$, and  $\mathcal{D}^{\mathrm{OE}}_{\mathrm{out}}$. Samples from $\mathcal{D}_{\mathrm{in}}$ are in-distribution (ID). Samples from other distributions are OOD. 
$\mathcal{D}_{\mathrm{out}}$ is an {\em unknown}  distribution for OOD samples, which we will encounter at test time. $\mathcal{D}^{\mathrm{OE}}_{\mathrm{out}}$ is a {\em known} distribution of OOD samples, but it is unknown how similar to or different from $\mathcal{D}_{\mathrm{out}}$ it is.
A particular dataset is assumed to be available for $\mathcal{D}^{\mathrm{OE}}_{\mathrm{out}}$ and utilized for training the model. 
The idea of the method is to train a model using samples from both $\mathcal{D}_{\mathrm{in}}$ and $\mathcal{D}^{\mathrm{OE}}_{\mathrm{out}}$ so that the model classifies the $\mathcal{D}_{\mathrm{in}}$ samples correctly, whereas it predicts a uniform probability distribution for all the classes for $\mathcal{D}^{\mathrm{OE}}_{\mathrm{out}}$ samples. Specifically, denoting the model by $\hat{y}=f(x)$ for $K$-class classification,  the method minimizes the following loss: 
\begin{equation}
   \mathbb{E}_{(x,y)\sim\mathcal{D}_{\mathrm{in}}}[\mathrm{CE}(f(x),y)] + \lambda \mathbb{E}_{x\sim\mathcal{D}^{\mathrm{OE}}_{\mathrm{out}}}[\mathrm{CE}(f(x),\mathcal{U}_K)],
    \label{eqn:oecost}
\end{equation}
where $\mathrm{CE}$ is cross-entropy; $y$ is the target distribution represented as a hard label (i.e., a one-hot vector) of the true class;
$\mathcal{U}_K$ represents uniform distribution over $K$ classes, i.e.,  $y=[1/K,1/K,\cdots,1/K]$.
The method aims to ``learn a more conservative concept of the ID samples and enable the detection of novel forms of anomalies''. 

Their paper reports experimental results that show good performance of the method. The major issue with the method is that it is unclear how to specify $\mathcal{D}^{\mathrm{OE}}_{\mathrm{out}}$ (or precisely a dataset for $\mathcal{D}^{\mathrm{OE}}_{\mathrm{out}}$). It easy to specify an arbitrary dataset, but it is not guaranteed to lead to good results. Although it is formally distinguished from  $\mathcal{D}_{\mathrm{out}}$ in their experiments, the result will inevitably depend on their similarity. It is hard to examine how their (dis)similarity affects the performance. 
Therefore, it remains unclear if the method works well for various problems in the real world. 

\subsection{Corrupted Images as OOD Samples}

It is ideal not to assume any specific OOD distribution or dataset. The problem is how to achieve good performance without it. An approach is to synthesize OOD samples from ID samples. Indeed, several previous studies \cite{lee2018simple,zisselman2020resflow} propose to create adversarial examples from ID samples and use them as imaginary OOD samples. However, they only use the created samples to adjust hyperparameters to maximize OOD detection performance; the adversarial examples are not used for the training of networks.

\begin{figure}[thb]
    \centering
    \includegraphics[width=0.9\linewidth]{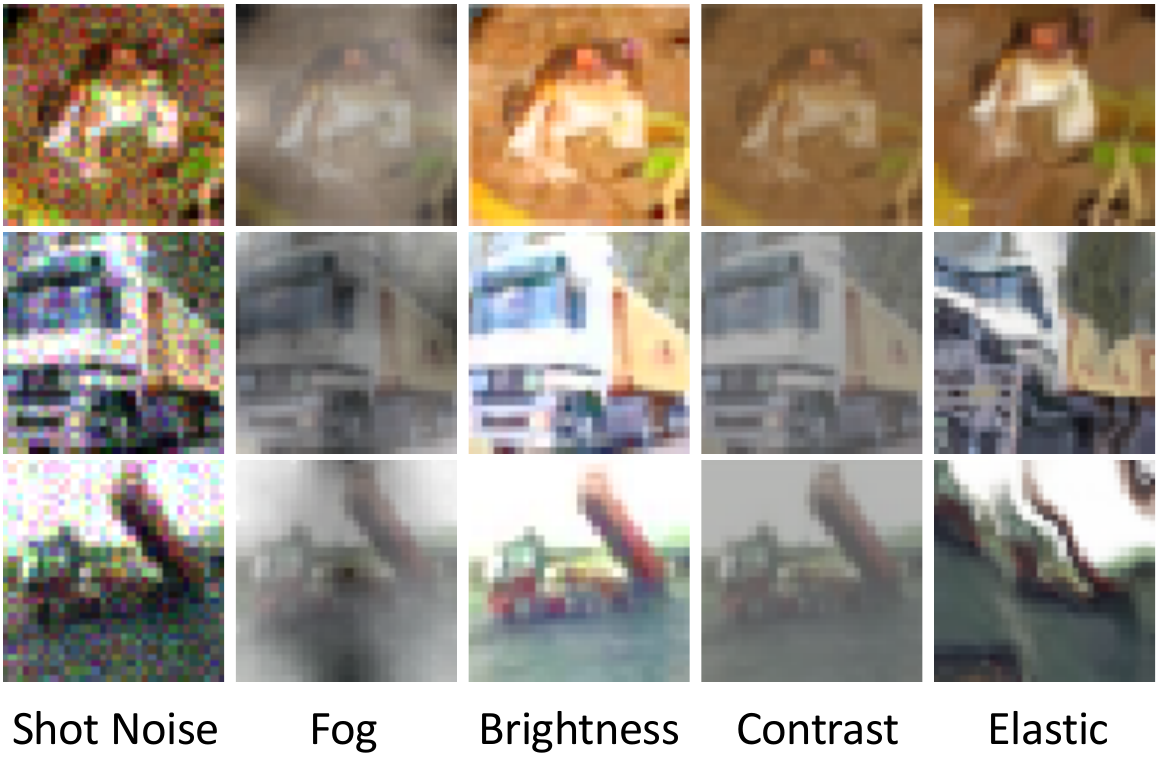}
    \caption{Examples of corrupted images. Images from CIFAR-10 are corrupted by the method of \cite{hendrycks2019robustness}. }
    \label{fig::corrupt_example}
\end{figure}

Instead, we consider using corrupting and distorting images to synthesize OOD samples. Specifically, we corrupt ID samples in various ways, as shown in Fig.~\ref{fig::corrupt_example}, and regard the corrupted images as OOD samples. The underlying thought is that severely corrupted ID images will become OOD samples.

Such image corruption has been considered a data augmentation method, where the corrupted images are treated as ID samples. It is widely recognized \cite{hendrycks2019robustness} that CNNs trained only on clean images tend to fail to classify corrupted images correctly. Researchers have paid attention to generalizing the models to such image corruption \cite{geirhos2018robustness,hendrycks2020augmix,rusak2020simple}. It is then natural to use image corruption as a data augmentation method, requiring the created data to maintain the original samples' semantic contents.

To test the idea of using corrupted images as OOD samples, we conducted preliminary experiments. Concretely, we use the method to create the  
ImageNet-C dataset \cite{hendrycks2019robustness}, which is publicly available by the authors\footnote{https://github.com/hendrycks/robustness} to synthesize various types of image corruption. It can synthesize 19 types of corruption, for each of which we can specify 5 severity levels.
We choose 15 out of 19 corruption types that are originally assumed for training uses.

Choosing CIFAR-100 for ID samples, we train a network (i.e., Wide Resnet 40-4) in the following four settings. The first is to train the model using only the original, clean images. We apply the standard data augmentation (i.e., random crop and horizontal flip), which is also the case with the rest of the three.
The second is train the model with the corrupted images, where image corruption is treated as  data augmentation; in other words, the model is trained so as to classify the corrupted images to their original ground truth classes. 
The third is to train the model with 1:1 population of clean images and corrupted images, in which the clean images are treated as ID samples and the corrupted images are treated as OOD samples, and then the loss (\ref{eqn:oecost}) is minimized as in Outlier Exposure (OE). The image corruption employed in the last two settings is randomly chosen from the 5 severity levels of the 15 corruptions. For the sake of comparison, we also consider another setting for OE, where we create adversarial examples from ID samples and use them as OOD samples, as in done in many studies on OOD detection \cite{lee2018simple,zisselman2020resflow}.

\begin{figure}[tb]
    \centering
    \includegraphics[width=1.\linewidth]{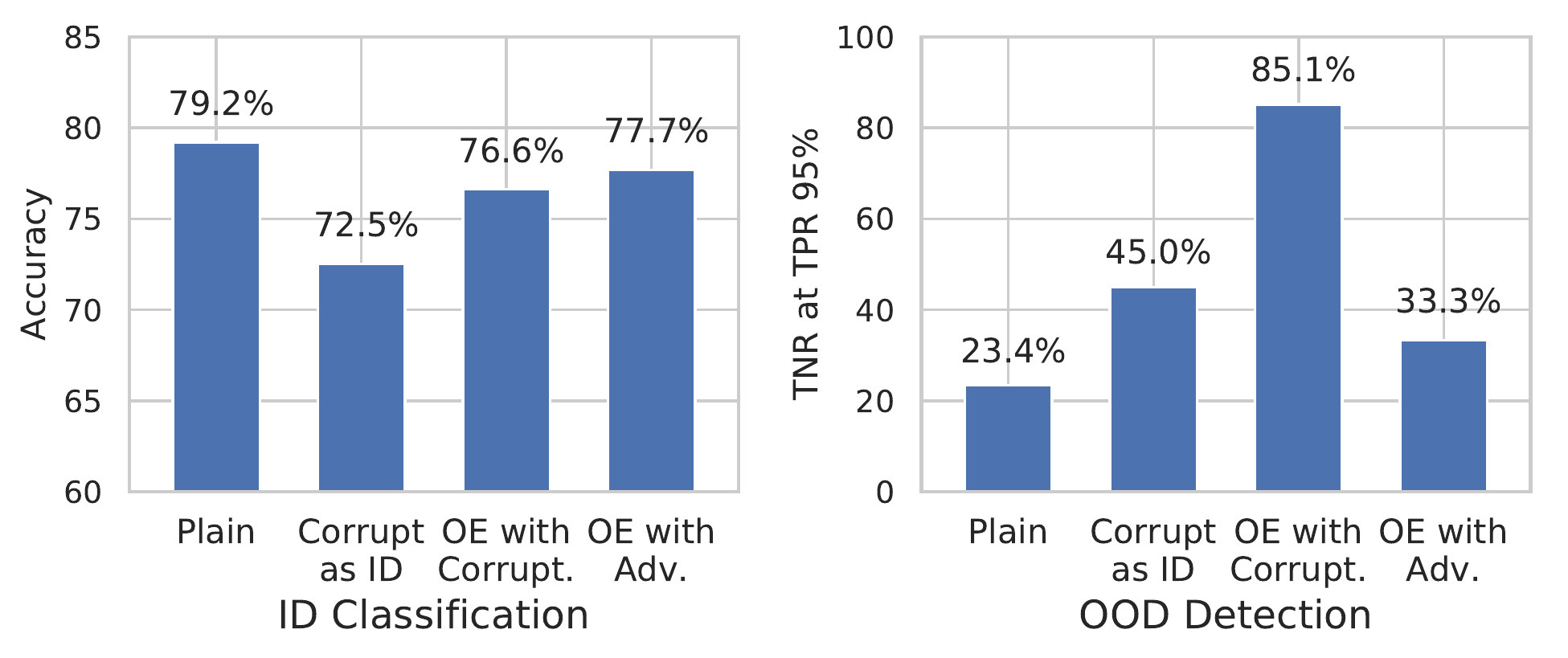}
    \caption{Comparison between classification and OOD detection performance of three models. "Plain" indicates the model trained with the standard training, while "Corrupt as ID" and "Corrupt as OOD" apply the ImageNet-C corruptions to the training dataset and utilize them as ID and OOD, respectively. 
    OOD detection performance is evaluated by \emph{TNR at TPR 95\%}.}
    \label{fig::plain_vs_corrupt}
\end{figure}

We tested the four models in terms of ID classification accuracy and OOD detection performance. We assumed the standard datasets (i.e., CIFAR-10, TIN, LSUN, iSUN, SVHN, and Food-101) for OOD, following previous studies; see Sec.~\ref{sec::experiment_settings} for details of the experiments. 
Figure~\ref{fig::plain_vs_corrupt} shows the results.
It is observed that the two models trained with corrupted images lose the ID classification accuracy  whereas they achieve better OOD detection performance. 

These results have several implications. On the one hand, we can confirm some of the corrupted images do work as OOD samples as we intend, as the third model (i.e., OE using the corrupted images) improves OOD detection performance. This is also supported by the fact that the second model (i.e., the one trained using corrupted images as augmented data) loses ID classification accuracy. This performance deterioration is understandable because the second model uses all the corrupted images as ID samples, but some do not preserve the original semantic contents due to their severity; learning them as ID samples will harm the classification accuracy. On the other hand, some of the corrupted images work as ID samples; those with mild corruption levels preserve the image contents. This is verifiable by the fact that the third model loses ID classification accuracy (i.e., 79.2\% $\rightarrow$ 76.6\%). It is also noted that the last model using adversarial examples as OOD does not perform well on OOD detection compared with the third model using corrupted images as OOD samples.

Based on the above results and discussion, we pose the following conjectures: 
\begin{itemize}
\item First, we should treat some of the corrupted images as ID samples and some of them as OOD samples. It will not be wise to treat all of them as either ID or OOD as we do in the above experiments. 
\item The employed image corruption will continuously cover the spectrum between ID and OOD samples due to the five severity levels of 15 corruptions.
\end{itemize}
These suggest that we could make further improvements by considering the {\em intermediate} region between ID and OOD. If we can assign intermediate labels to them, we could extend the OOD detection performance while maintaining the ID classification accuracy.

\subsection{Soft Labels for Intermediate between ID and OOD}
\label{sec::train_soft_target}

The question is how to assign soft labels to the intermediate samples between ID and OOD. Our solution is based on two ideas. One is to use a model $f(x)$ trained on the ID classification task using $\mathcal{D}_{\mathrm{in}}$ to obtain the soft labels. 
The other is to assign a soft label to a corrupted image $x'=T(x)$ according to the image corrupting transformation $T$, which is one of the predefined transformations. This is  a natural extension of the above finding. 

We denote an image corrupting transformation by $x'=T_i(x)$, where $i(=1,\ldots,75(=15\times 5))$ is an index indicating one of the 15 image corruption types with five severity levels. 
We apply $T_i$ to each sample of $\mathcal{D}_{\mathrm{in}}$. Let $\mathcal{D}^*_{\mathrm{in},i}$ be the set of the corrupted images. Making the model $f(x)$ trained with clean images classify each sample of $\mathcal{D}^*_{\mathrm{in},i}$, we calculate the classification accuracy $\mathrm{acc}_i$ as
\begin{equation}
\mathrm{acc}_i=\frac{1}{N}\sum_{n=1}^N 1(\hat{k}_n=k_n),
\end{equation}
where $N=|\mathcal{D}_{\mathrm{in}}|=|\mathcal{D}^*_{\mathrm{in},i}|$; $\hat{k}_n$ and $k_n$ are the predicted and true class indexes, respectively. We use  $\mathrm{acc}_i$ for obtaining a soft label as explained below.

\begin{figure}[tb]
    \centering
    \includegraphics[width=1.\linewidth]{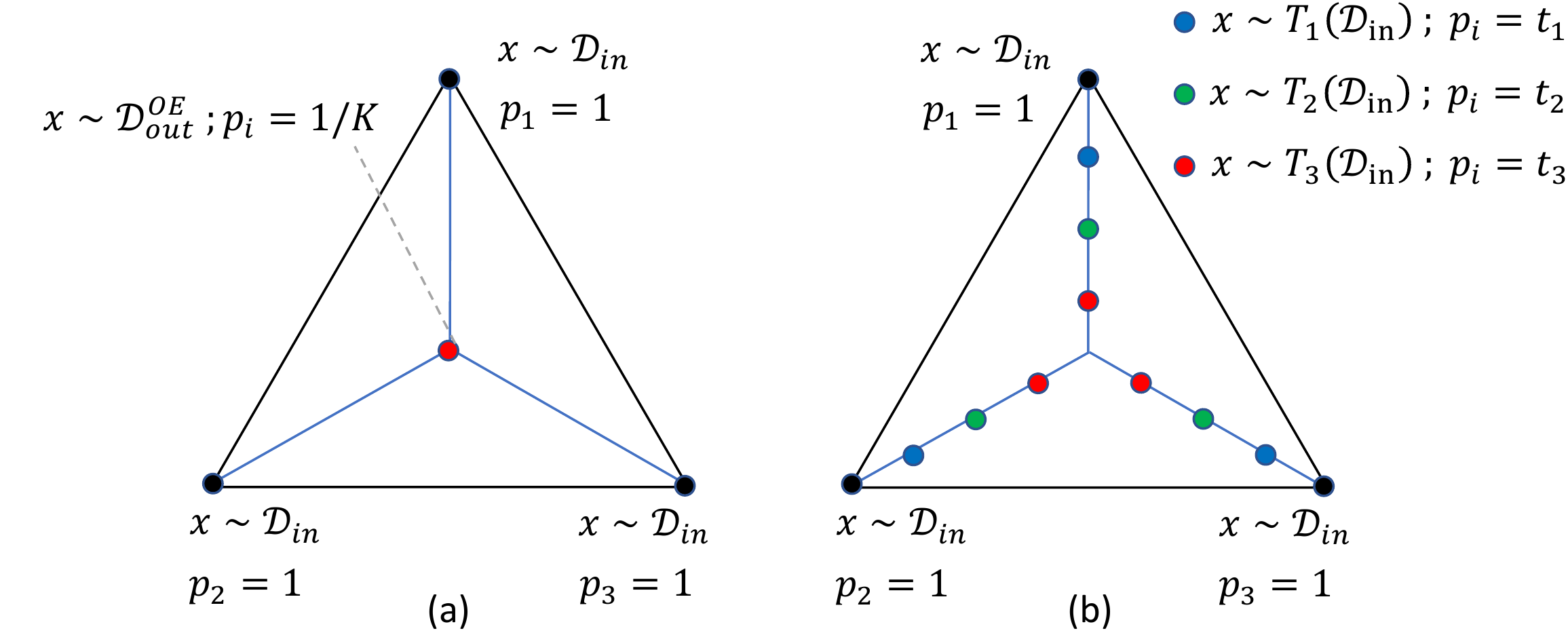} 
    \caption{Illustration of how target labels are assigned to ID and OOD samples in the case of three-class classification for (a) Outlier Exposure (OE) \cite{hendrycks2018outlier_oe} and (b) our method. In OE, the target class probabilities for ID samples are $1$ for the true class and $0$ for others; those for OOD are all $1/3$. In our method, there are intermediate samples lying between ID and OOD, which are created by applying image corrupting transformation $T_i$ to ID samples. Their target labels are set to constants determined for each $T_i$.}
    \label{fig::proposed_figure}
\end{figure}

We train a new model $f^*$ using $\mathcal{D}_{\mathrm{in}}$ and their corrupted images in the following way. Choosing a sample $x$ from $\mathcal{D}_{\mathrm{in}}$, we first decide whether we apply image corruption to it according to a probability $\gamma$. If not, we use the original hard label $y$ 
and employ the loss $- \log f_k(x)$, 
where $k$ is the true class index.
If yes, we choose and apply a transformation $T_i$ to $x$ and obtain a corrupted image $x'=T_i(x)$. We set the soft target label $t\in\mathbb{R}^K$ for $x'$ as follows:
\begin{equation} \label{eqn::produce_soft_label}
    t_j = \left\{
    \begin{array}{ll}
    \mathrm{acc}_i & \mbox{if $j=k$},\\
    \frac{1-\mathrm{acc}_i}{K-1} & \mbox{otherwise},
    \end{array}\right.
\end{equation}
where $j=1,\ldots,K$ and $k$ is the true class index of $x$.
This soft label is considered to be an interpolation between the hard label for the true class and the uniform label for OOD, as shown in Fig.~\ref{fig::proposed_figure}.

When choosing $T_i$ for each $x\in\mathcal{D}_{\mathrm{in}}$, we choose one so that the soft label $t$ distributes as uniformly in the label space as possible over $\mathcal{D}_{\mathrm{in}}$. Rigorously, the maximum element of $t$ (i.e., $\mathrm{argmax}_l ~t_l$), which is given by $\mathrm{acc}_i$ as in (\ref{eqn::produce_soft_label}), distributes uniformly in the range $[1/K,1]$. To do this, we sample a random number $\alpha$ from a uniform distribution of the range $[1/K,1]$ and search the nearest neighbor $\mathrm{acc}_i$ to $\alpha$ from the pre-computed set $\{\mathrm{acc}_i\}_{i=1,\ldots,75}$ and choose the corresponding transformation $T_i$. 

The procedure of creating a sample $(x,y)$ with or without image corruption from $\mathcal{D}_{\mathrm{in}}$ is summarized in Algorithm~\ref{alg::draw_corrupt_sample}; $(x,y)$ is either the pair of an original image and a hard label or the pair of a corrupted image and a soft label (i.e., $y_j\leftarrow t_j$ for $j=1,\ldots,K$). To train $f^*$, we minimize the cross entropy loss between a sample $(x,y)$ created as above and the prediction $\hat{y}=f^*(x)$:
\begin{equation}
     \mathrm{CE}(f^*(x),y) = - \sum_{j=1}^K y_j \log f_j^*(x).
\end{equation}

\begin{algorithm}[th]
\label{alg::draw_corrupt_sample}
\SetAlgoLined
\textbf{Input:} Dataset $\mathcal{D}$, probability $\gamma$ of applying image corruption, and $\{\mathrm{acc}_i\}_{i=1,\ldots,75}$\\
\textbf{Output:} An input image $x$ and its label $y$ \\
  Sample $(x,y)$ from $\mathcal{D}$\;
  $\beta\sim\mathrm{Bernouli(\gamma)}$\;
  \If{$\beta = 1$}{
    $\alpha \sim \mathcal{U}(1/K,1)$\;
    $i\leftarrow \mathrm{argmin}_j |\mathrm{acc}_j - \alpha|$\;
    $x\leftarrow T_i(x)$\;
    Set $t$ according to (\ref{eqn::produce_soft_label})\;
    $y\leftarrow t$\;
  }
 \Return $(x,y)$
 \caption{Obtaining a training sample $(x,y)$}
\end{algorithm}

There are two choices with how to train $f^*$. One is to initialize $f^*\leftarrow f$, where $f$ is the model trained using $\mathcal{D}_{\mathrm{in}}$ on the ID class classification. The other is to train $f^*$ from scratch. We found in our experiments that the latter consistently works better and thus we will report its performance.

Previous studies employ 
the maximum softmax probability (MSP), i.e., $\max_l f_l(x)$, or its variant for detecting OOD samples \cite{hendrycks2016baseline,liang2017enhancing}. Our experiments show that the predicted  entropy $-\sum_{j=1}^K f_j(x) \log f_j(x)$ shows slightly better performance consistently.
Thus, we will report the results obtained using the entropy. Specifically, we classify an input providing the entropy higher than a threshold as OOD. 

In our method, the probability $\gamma$ of applying image corrupting transformation to each sample is a hyperparameter. As will be shown later, the performance of the proposed method is not sensitive to its $\gamma$; it is stable in the range $[0.01,0.4]$. We set $\gamma=0.2$ in our experiments based on the ID classification performance on the validation data.

\section{Experiments} \label{sec::experiments}

We conduct experiments to evaluate the proposed method and compare it with existing methods. 

\subsection{Experimental Settings} \label{sec::experiment_settings}

We consider an image classification task, for which its training dataset $\mathcal{D}_{\mathrm{in}}$ is given. Each OOD detection method first trains a network $f$ so that $f$ will accurately classify an ID input image $x$ as $\hat{y}=f(x)$. Then the method judges whether a new input $x$ is ID or OOD. We evaluate how accurately it can detect OOD samples.

\medskip
\noindent\textbf{Datasets}~~ In the experiments, we choose four ID datasets, i.e., CIFAR-10/100 \cite{krizhevsky2009cifar}, SVHN \cite{netzer2011svhn}, and Food-101 \cite{bossard14food101}. We use one of the four datasets for ID and treat all others for OOD. Following Liang et al. \cite{liang2017enhancing}, we also use Tiny ImageNet \cite{deng2009imagenet} (cropped and resized), LSUN \cite{yu15lsun} (cropped and resized), and iSUN \cite{xu2015isun} for OOD. CIFAR-10 and 100 \cite{krizhevsky2009cifar} are datasets of 10 and 100 object categories, respectively, and each contains 50,000 and 10,000 samples for training and testing, respectively. SVHN \cite{netzer2011svhn} is a dataset of digit classification containing of 73,257 and 26,032 samples for training and testing. Food-101 \cite{bossard14food101} is a dataset of 101 food categories with 75,750 and 25,250 samples for training and testing. We resize the images of Food-101 to $32 \times 32$ pixels; the images of other datasets are of the same size.

\medskip
\noindent\textbf{Networks and Training}~~ We use DenseNet \cite{huang2017densely} and Wide Resnet \cite{zagoruyko2016wideresnet}. More specifically, we use a 100-layer DenseNet with the bottleneck block and the growth rate of $12$, denoted by DenseNet-100-12. We train it on an ID dataset for 300 epochs with mini-batch size of 64. For Wide Resnet, we use the 40-layer Wide Resnet with a widen factor of 4, denoted by WRN-40-4. We train it for 200 epochs with mini-batch size of 128. 
We employ weight regularization with factor 0.0001 for the former and with 0.0005 for the latter. 

\medskip
\noindent\textbf{Evaluation Metrics}~~
Following previous studies \cite{hendrycks2016baseline,liang2017enhancing,lee2018simple,Yu2019UnsupervisedOD,yu2020compression,sastry2020gram_matrices}, we employ   three standard metrics for evaluating OOD detection performance, i.e., {\em true negative rate at true positive rate 95\% (TNR at TPR 95\%)}, area under the ROC curve (AUROC), and area under the precision-recall curve (AUPR). 
Note that for all these metrics, a higher value indicates better performance.

\subsection{Compared Methods} \label{sec::compared_methods}

We compare our method with the following five methods. 
Following the recent studies \cite{shafaei2018biased,Yu2019UnsupervisedOD,techapanurak2019hyper,hsu2020generalize,yu2020compression,sastry2020gram_matrices}, we confine ourselves to the methods that do not need explicit OOD samples for training, if they are only a few. An exception is Outlier Exposure \cite{hendrycks2018outlier_oe}, which uses OOD samples for training but does not assume the similarity between them and the real OOD samples we encounter at test time.

\medskip
\noindent\textbf{Baseline Method} \cite{hendrycks2016baseline}\quad This method trains the network using $\mathcal{D}_{\mathrm{in}}$ in the standard way. We call the resulting model the vanilla model. The method threshold the maximum softmax probability (MSP), also known as confidence, provided by the vanilla model for an input $x$ to judge if it is OOD. We use the predicted entropy from the same output in our experiments, because we found that it yields slightly better performance than MSP. 

\medskip
\noindent\textbf{Cosine Similarity} \cite{hsu2020generalize}\quad This method uses a scaled cosine similarity instead of the dot product to yield logits in the final layer of the networks. Thus, it needs to change the layer design. It also employs a variant of input perturbation  \cite{liang2017enhancing} for improved detection that is feasible without explicit OOD samples.

\medskip
\noindent\textbf{Gram Matrices} \cite{sastry2020gram_matrices}\quad This method utilizes the Gram matrix calculated from 
the intermediate layer features of the vanilla model. 
It learns its statistics from $\mathcal{D}_{\mathrm{in}}$ and use it to detect OOD inputs. 

\medskip
\noindent\textbf{MALCOM} \cite{yu2020compression}\quad This method extends the Mahalanobis detector by the compression distance. The feature vectors are extracted both from the global average pooling (GAP) and the compression complexity pooling (CCP). All vectors are combined through the concatenation and are used to model the Mahalanobis distance.

\medskip
\noindent\textbf{Outlier Exposure} \cite{hendrycks2018outlier_oe}\quad 
We have explained this method in Sec.~\ref{sec::revisiting_outlier_exposure}. Following the study, we use 80 Million Tiny Images datasets \cite{torralba200880m_tiny_image} for $\mathcal{D}^{\mathrm{OE}}_{\mathrm{out}}$ in our experiments. We use the entropy instead of MSP due to the same reason as above.

\subsection{Results} \label{sec::experimental_results}

\subsubsection{Out-of-Distribution Detection} \label{sec::ood_detection_result}

We evaluate the OOD detection performance of the above methods for each ID dataset. Tables \ref{table::overall_ood_results_tnr} and \ref{table::overall_ood_results_auroc} show the performance measured by TNR at TPR 95\% and AUROC, respectively. They show the mean and standard deviation over all the OOD datasets and over five trials of training. Those measured by AUPR and the detailed results showing the performance on each OOD dataset separately are given in the supplementary material.

It is seen that the proposed method achieves the best performance in almost all cases and in any evaluation metrics. We treat Outlier Exposure separately, as it uses external datasets; its performance should depend on their choice, although they are different from the true OOD datasets in our experiments. Nevertheless, our method performs comparably well and yields better results in several cases, i.e., when ID is CIFAR-100 and Food-101. 

\begin{table*}[tbh]
    \caption{The OOD detection accuracy of the compared methods measured by TNR at TPR 95\%. Outlier Exposure (OE) is treated separately, as its performance depends on the OOD dataset it assumes.  }
    \label{table::overall_ood_results_tnr}
    \begin{center}
    \begin{small}
    \begin{tabular}{@{\hskip 0.1in}c@{\hskip 0.1in}c@{\hskip 0.1in}c@{\hskip 0.1in}c@{\hskip 0.1in}c@{\hskip 0.1in}c@{\hskip 0.1in}c@{\hskip 0.1in}|c@{\hskip 0.1in}}
    \toprule
    Net & ID & Baseline & Cosine & Gram & MALCOM & Ours & OE \\
    \midrule
    \multirow{4}{*}{\rotatebox[origin=c]{90}{DenseNet}} 
    & CIFAR-10 & 51.61(9.71) & 88.07(13.73) & 78.28(31.52) & 74.64(30.39) & \B91.59(17.19) & \B93.83(8.65) \\
    & CIFAR-100 & 27.47(7.45) & 80.46(20.87) & 68.80(34.01) & 56.65(32.82) & \B85.69(27.83) & 51.81(15.93) \\
    & SVHN & 68.53(4.40) & 74.27(10.70) & 93.39(6.69) & 98.70(1.50) & \B99.32(0.82) & \B99.92(0.15) \\
    & Food-101 & 12.45(4.44) & 88.62(7.82) & 73.68(27.46) & 82.15(22.97) & \B89.71(8.80) & 63.46(28.98) \\
    \midrule
    \multirow{4}{*}{\rotatebox[origin=c]{90}{WRN}} 
    & CIFAR-10 & 53.78(7.29) & 81.53(15.40) & 80.90(29.06) & 80.58(25.89) & \B92.05(16.86) & \B95.28(6.51) \\
    & CIFAR-100 & 25.19(7.05) & 67.06(16.22) & 70.61(32.89) & 63.97(28.39) & \B86.04(27.33) & 42.82(17.18) \\
    & SVHN & 73.35(3.27) & 75.99(11.31) & 94.00(6.35) & 98.42(1.66) & \B99.44(0.69) & \B99.97(0.08) \\
    & Food-101 & 11.77(3.05) & 77.47(8.94) & 75.62(27.67) & 82.88(22.90) & \B85.44(14.51) & 80.02(17.49) \\
    \bottomrule
    \end{tabular}
    \end{small}
    \end{center}
\end{table*}

\begin{table*}[tbh]
    \caption{The OOD detection accuracy of the compared methods measured by AUROC.}
    \label{table::overall_ood_results_auroc}
    \begin{center}
    \begin{small}
    \begin{tabular}{@{\hskip 0.1in}c@{\hskip 0.1in}c@{\hskip 0.1in}c@{\hskip 0.1in}c@{\hskip 0.1in}c@{\hskip 0.1in}c@{\hskip 0.1in}c@{\hskip 0.1in}|c@{\hskip 0.1in}}
    \toprule
    Net & ID & Baseline & Cosine & Gram & MALCOM & Ours & OE \\
    \midrule
    \multirow{4}{*}{\rotatebox[origin=c]{90}{DenseNet}} 
    & CIFAR-10 & 92.08(3.53) & 97.29(3.23) & 92.56(11.48) & 92.80(9.99) & \B97.86(4.52) & \B98.54(1.69) \\
    & CIFAR-100 & 80.00(4.04) & 95.30(6.43) & 89.78(13.23) & 86.23(17.49) & \B96.10(8.22) & 88.91(4.53) \\
    & SVHN & 92.05(2.15) & 94.06(2.76) & 98.44(1.56) & 99.57(0.40) & \B99.79(0.21) & \B99.98(0.04) \\
    & Food-101 & 64.36(4.91) & \B97.82(1.45) & 91.66(10.16) & 95.46(5.90) & 97.00(2.69) & 87.43(10.78) \\
    \midrule
    \multirow{4}{*}{\rotatebox[origin=c]{90}{WRN}} 
    & CIFAR-10 & 90.96(2.64) & 95.98(3.68) & 94.76(8.21) & 95.64(5.98) & \B97.81(4.67) & \B98.36(1.06) \\
    & CIFAR-100 & 78.05(4.38) & 92.81(5.80) & 91.34(11.25) & 91.80(8.73) & \B95.82(8.98) & 89.51(4.43) \\
    & SVHN & 93.06(1.29) & 94.56(2.76) & 98.61(1.43) & 99.58(0.41) & \B99.78(0.16) & \B99.98(0.02) \\
    & Food-101 & 66.46(2.59) & \B95.84(1.71) & 92.54(9.41) & 95.53(6.16) & \B95.85(4.24) & 93.10(6.18) \\
    \bottomrule
    \end{tabular}
    \end{small}
    \end{center}
\end{table*}

\subsubsection{In-Distribution Classification } \label{sec::id_classification}

\begin{table}[tbh]
    \caption{The ID classification performance.}
    \label{table::id_classification_results}
    \begin{center}
    \begin{small}
    \resizebox{1.0\linewidth}{!}{
    \begin{tabular}{@{\hskip 0.1in}c@{\hskip 0.1in}c@{\hskip 0.1in}c@{\hskip 0.1in}c@{\hskip 0.1in}c@{\hskip 0.1in}|c@{\hskip 0.1in}}
    \toprule
    & ID & Standard & Cosine & Ours & OE \\
    \midrule
    \multirow{4}{*}{\rotatebox[origin=c]{90}{DenseNet}} 
    & CIFAR-10 & 95.13(0.09) & 94.89(0.13) & \B95.37(0.12) & 94.99(0.09) \\
    & CIFAR-100 & 76.94(0.37) & 75.39(0.50) & \B77.71(0.17) & 75.93(0.33) \\
    & SVHN & 96.36(0.10) & 95.98(0.18) & \B96.63(0.04) & 96.49(0.06) \\
    & Food-101 & 42.42(0.23) & 40.53(0.24) & \B42.88(0.36) & \B46.88(0.23) \\
    \midrule
    \multirow{4}{*}{\rotatebox[origin=c]{90}{WRN}} 
    & CIFAR-10 & 95.54(0.13) & 95.10(0.18) & \B95.59(0.12) & \B95.79(0.10) \\
    & CIFAR-100 & \B79.28(0.30) & 76.66(0.30) & 79.21(0.11) & 76.58(0.23) \\
    & SVHN & 96.67(0.03) & 96.40(0.10) & \B96.83(0.06) & 96.69(0.04) \\
    & Food-101 & \B44.92(0.15) & 43.22(0.19) & 44.76(0.20) & \B45.00(0.23) \\
    \bottomrule
    \end{tabular}
    }
    \end{small}
    \end{center}
\end{table}

Some of the compared methods change either the network architecture or its training method \cite{hendrycks2018outlier_oe,techapanurak2019hyper,hsu2020generalize} for OOD detection. These changes sometimes result in lower ID classification accuracy than the original network trained in the standard fashion. 
Table~\ref{table::id_classification_results} shows the ID classification accuracy for the standard model, the scaled cosine similarity, the proposed method, and Outlier Exposure. It is observed that the proposed method yields comparable performance to the standard model, whereas the cosine similarity underperforms slightly.
All the other compared methods i.e., Gram Matrices \cite{sastry2020gram_matrices} and MALCOM \cite{yu2020compression}, use the standard model.

\subsubsection{Calibration Errors} \label{sec::model_calibration}

It is well recognized \cite{guo2017calibration} that modern neural networks tend to be over-confident with their prediction for classification tasks. Specifically, when they classify an input, the confidence of the prediction (i.e., the maximum softmax probability) tends to be larger than the expected prediction accuracy. It is said to be ``calibrated'' when the two are well aligned. How well a model is calibrated is evaluated by the expected calibration error (ECE). We calculate ECE for the model trained in each method. Table~\ref{table::calibration_result} shows the results. It is seen that the proposed method achieves the smallest ECE in most cases. 

\begin{table}[tbh]
    \caption{Calibration errors of the models of the compared methods measured by the expected calibration error (ECE). A lower value is better.}
    \label{table::calibration_result}
    \begin{center}
    \begin{small}
    \resizebox{1.0\linewidth}{!}{
    \begin{tabular}{@{\hskip 0.1in}c@{\hskip 0.1in}c@{\hskip 0.1in}c@{\hskip 0.1in}c@{\hskip 0.1in}c@{\hskip 0.1in}|c@{\hskip 0.1in}}
    \toprule
    Net & ID & Standard & Cosine & Ours & OE \\
    \midrule
    \multirow{4}{*}{\rotatebox[origin=c]{90}{DenseNet}} 
    & CIFAR-10 & 2.93(0.13) & 4.30(0.14) & \B1.01(0.21) & 2.07(0.15) \\
    & CIFAR-100 & 12.12(0.44) & 20.48(0.61) & \B3.63(0.44) & 4.51(0.16) \\
    & SVHN & 2.39(0.05) & 3.04(0.15) & \B0.90(0.14) & 1.00(0.09) \\
    & Food-101 & 33.85(0.36) & 44.84(1.09) & \B25.03(1.78) & \B10.39(0.49) \\
    \midrule
    \multirow{4}{*}{\rotatebox[origin=c]{90}{WRN}} 
    & CIFAR-10 & 2.86(0.13) & 4.14(0.10) & \B0.58(0.08) & 6.29(0.10) \\
    & CIFAR-100 & 10.68(0.17) & 19.12(0.26) & \B4.76(0.10) & 17.24(0.28) \\
    & SVHN & 2.18(0.07) & 2.93(0.11) & \B1.56(0.55) & 1.91(0.02) \\
    & Food-101 & 28.34(0.18) & 47.59(0.54) & \B19.26(1.41) & 32.09(0.24) \\
    \bottomrule
    \end{tabular}
    }
    \end{small}
    \end{center}
\end{table}

\subsection{Analyses} \label{sec::ablation_study}

\subsubsection{What Image Corrupting Transformation Is Good?}

In the above experiments, we used 15 image corrupting transformations. The following questions will arise. Which corrupting transformations are more effective? How many transformations are necessary? To analyze these, we examine how the results will change depending on the number of transformations or a particular combination of selected corrupting transformation. We choose CIFAR-100 for the ID dataset and employ the Wide Resnet, which corresponds to a row in Table \ref{table::overall_ood_results_tnr}. 

Table~\ref{table::ablation_corruptions} shows OOD detection performance obtained for selected combinations of one to five corrupting transformations. The number(s) in $\{\cdot\}$ indicates the index of the 15 corruption types; see the supplementary material for their details.
Roughly speaking, using a combination of more corruption types tends to yield better performance.
That said, some combinations of a few corruption types work much better than others. If we choose only a single corruption type, the best performer is \emph{Elastic transform} (denoted by \{12\} in the table); its performance (i.e., $77.88$) is still better than other existing methods; see Table \ref{table::overall_ood_results_tnr}. Overall, we suggest to use an ensemble of all the corruption types, as it yields the best performance in the combinations we tested.

\begin{table}[tbh]
    \caption{The OOD detection result in TNR at TPR 95\% utilizing different set of the image-corrupting transformatoins in the training.}
    \label{table::ablation_corruptions}
    \begin{center}
    \begin{small}
    \begin{tabular}{@{\hskip 0.1in}c@{\hskip 0.2in}c@{\hskip 0.1in}}
        \begin{tabular}{@{\hskip 0.1in}c@{\hskip 0.1in}c@{\hskip 0.1in}c@{\hskip 0.1in}c@{\hskip 0.1in}}
        \toprule
        Corruption & TNR \\
        \midrule
        \{0\} & 37.20 \\
        \{1\} & 39.19 \\
        \{2\} & 25.87 \\
        \{3\} & 34.94 \\
        \{4\} & 37.20 \\
        \{5\} & 37.88 \\
        \{6\} & 32.79 \\
        \{7\} & 26.89 \\
        \{8\} & 34.90 \\
        \{9\} & 30.44 \\
        \{10\} & 30.23 \\
        \{11\} & 26.75 \\
        \bottomrule
        \end{tabular}
    &
        \begin{tabular}{@{\hskip 0.1in}c@{\hskip 0.1in}c@{\hskip 0.1in}}
        \toprule
        Corruption & TNR  \\
        \midrule
        \{12\} & 77.88 \\
        \{13\} & 31.22 \\
        \{14\} & 62.98 \\
        \{0, 1, 2\} & 59.70 \\
        \{3, 4, 5\} & 48.65 \\
        \{6, 7, 8\} & 76.84 \\
        \{9, 10, 11\} & 64.98 \\
        \{12, 13, 14\} & 58.44 \\
        \{0, 1, 2, 3, 4\} & 78.80 \\
        \{5, 6, 7, 8, 9\} & 63.79 \\
        \{10, 11, 12, 13, 14\} & 83.06 \\
        All & \B86.04 \\
        \bottomrule
        \end{tabular}
    \end{tabular}
    \end{small}
    \end{center}
\end{table}

\subsubsection{Sensitivity to $\gamma$}

\begin{figure}[tb]
    \centering
    \includegraphics[width=0.9\linewidth]{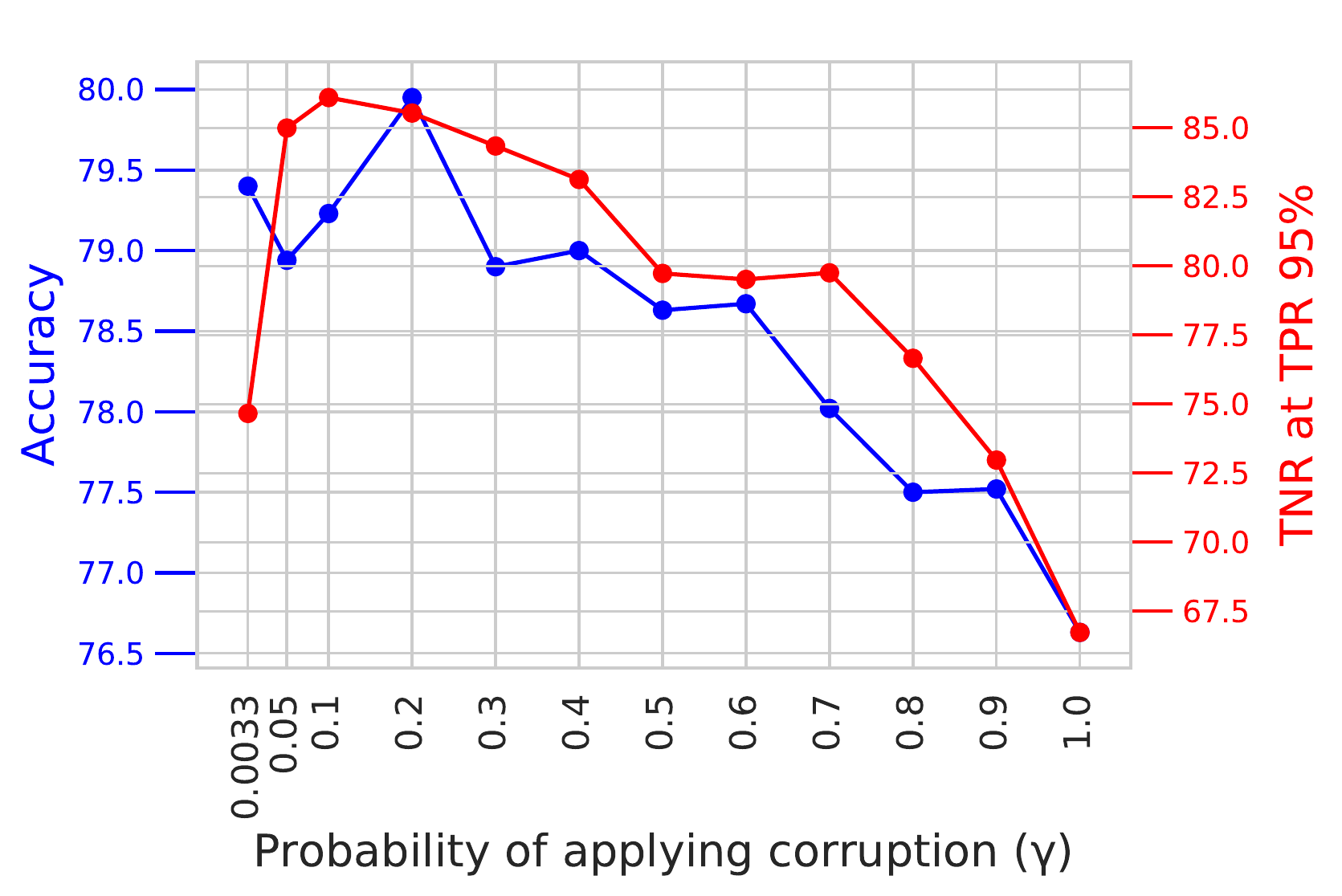}
    \caption{ID classification accuracy and OOD detection accuracy (TNR at TPR 95\%) vs. the probability $\gamma$ of applying corruption to each image.}
    \label{fig::choosing_prob}
\end{figure}

The proposed method has a hyperparameter $\gamma$, which is the probability of applying image corruption to each sample at training time. We evaluate the sensitivity of the results to its choice. Figure \ref{fig::choosing_prob} shows the ID classification accuracy and the OOD detection accuracy for different $\gamma$'s ranging in $[0,1]$. 
We can observe the following. First, fortunately, the results are not sensitive to $\gamma$, especially for the range of $[0,0.5]$. Second, they show similar tendencies; both decrease as $\gamma$ increases. From the result, it seems reasonable to choose it depending on the ID classification accuracy. Note that we cannot determine $\gamma$ based on the OOD detection accuracy, as it is not available without OOD samples. We chose $\gamma=0.2$ based on this consideration in the above experiments. The results do not change that much if we choose $\gamma=0.1$ or 0.3, due to the above insensitivity.

\section{Conclusion}

In this paper, we have presented a new method for OOD detection. Questioning the premise of previous studies that ID and OOD samples are separated distinctly, we consider samples lying in the intermediate of the two and use them for training a network. We generate such samples using multiple image transformations that corrupt input images in various ways and with different severity levels. Specifically, applying one of the image transformations to ID samples, we assign the generated samples a soft target label representing how distant they are from the ID region. We compute the distance by using a network trained on clean ID samples for classifying those samples; we calculate their mean classification accuracy and use it to create the soft label. We then train the same network from scratch using the original ID samples and the generated samples with soft labels. The trained network can classify ID samples accurately. We detect OOD samples by thresholding the entropy of the predicted softmax probability. The experimental results show that our method outperforms the previous state-of-the-art in the standard benchmark tests widely employed in previous studies. We have also analyzed the effect of the number and particular combinations of image corrupting transformations on the performance.

{\small
\bibliographystyle{ieee_fullname}
\bibliography{main}
}

\clearpage
\onecolumn
\appendix

\begin{center}
{\LARGE \bfseries
{Supplementary Material for 
``Bridging In- and Out-of-distribution Samples for Their Better Discriminability''}
}
\end{center}
\vspace{0.4cm}

\section{Additional Result in AUPR} \label{apx::result_aupr}

Table~\ref{table::overall_ood_results_aupr} shows the OOD detection accuracy measured by AUPR, which we omit from the main paper to save space. The same observation holds as the evaluation by TNR at TPR 95\% and by AUROC.

\begin{table*}[htb]
    \caption{The OOD detection accuracy of the compared methods measured by AUPR. }
    \label{table::overall_ood_results_aupr}
    \begin{center}
    \begin{small}
    \begin{tabular}{@{\hskip 0.1in}c@{\hskip 0.1in}c@{\hskip 0.1in}c@{\hskip 0.1in}c@{\hskip 0.1in}c@{\hskip 0.1in}c@{\hskip 0.1in}c@{\hskip 0.1in}|c@{\hskip 0.1in}}
    \toprule
    Net & ID & Baseline & Cosine & Gram & MALCOM & Ours & OE \\
    \midrule
    \multirow{4}{*}{\rotatebox[origin=c]{90}{DenseNet}} 
    & CIFAR-10 & 90.66(9.45) & 96.13(4.84) & 87.24(20.05) & 91.28(12.19) & \B97.36(5.01) & \B98.35(1.82) \\
    & CIFAR-100 & 79.30(6.30) & 94.06(7.69) & 85.78(17.96) & 85.42(17.82) & \B95.95(7.37) & 87.97(3.78) \\
    & SVHN & 94.55(3.95) & 96.71(2.40) & 96.56(3.69) & 99.83(0.14) & \B99.89(0.12) & \B99.99(0.02) \\
    & Food-101 & 80.22(5.76) & \B98.94(0.64) & 95.03(6.06) & 97.90(2.71) & 98.36(1.47) & 93.69(4.98) \\
    \midrule
    \multirow{4}{*}{\rotatebox[origin=c]{90}{WRN}} 
    & CIFAR-10 & 87.58(6.64) & 94.52(5.60) & 91.76(13.24) & 94.61(7.44) & \B97.27(5.51) & \B98.52(1.08) \\
    & CIFAR-100 & 75.40(5.49) & 91.47(6.77) & 88.40(14.21) & 91.25(8.93) & \B95.25(9.24) & 90.52(3.16) \\
    & SVHN & 95.44(1.58) & 97.15(1.73) & 97.08(3.30) & 99.83(0.16) & \B99.90(0.08) & \B99.99(0.01) \\
    & Food-101 & 81.62(6.23) & 97.88(1.11) & 95.67(5.48) & \B97.92(2.87) & 97.77(2.18) & 96.25(3.35) \\
    \bottomrule
    \end{tabular}
    \end{small}
    \end{center}
\end{table*}

\section{List of Image-Corrupting Transformations}

We used ImageNet-C module [12] for the image-corrupting transformations. We used the code publicly available at \url{https://github.com/hendrycks/robustness}.

\begin{table}[htb] 
    \caption{The OOD detection result in TNR at TPR 95\% utilizing different set of the corruption(s) in the training.}
    \label{table::ablation_corruption_list}
    \begin{center}
    \begin{small}
    \begin{tabular}{@{\hskip 0.1in}c@{\hskip 0.2in}c@{\hskip 0.4in}c@{\hskip 0.2in}c@{\hskip 0.1in}}
    \toprule
    \# & Corruption & \# & Corruption \\
    \midrule
    0 & Gaussian noise & 8 & Frost \\
    1 & Shot noise & 9 & Fog \\
    2 & Impulse noise & 10 & Brightness \\
    3 & Defocus blur & 11 & Contrast \\
    4 & Frosted glass blur & 12 & Elastic transform \\
    5 & Motion blur & 13 & Pixelate \\
    6 & Zoom blur & 14 & JPEG compression \\
    7 & Snow &  &  \\
    \bottomrule
    \end{tabular}
    \end{small}
    \end{center}
\end{table}

\clearpage
\section{Results with Individual OOD Datasets} \label{apx::full_result_table}

In the main paper, we report only the {\em average} of the detection scores for the multiple OOD datasets we employed in the experiments. Tables \ref{table::ood_detection_performance_dense} - \ref{table::ood_detection_performance_wrn_aupr} report 
individual detection scores (i.e., TNR at TPR 95\%, AUROC, and AUPR) for each OOD dataset. Each score is the average of five trials. 

\begin{table*}[tbh]
    \caption{The OOD detection performance in TNR at TPR 95\%. The network architecture is DenseNet-100-12.}
    \label{table::ood_detection_performance_dense}
    \begin{center}
    \begin{small}
    \begin{tabular}{@{\hskip 0.1in}c@{\hskip 0.1in}c@{\hskip 0.1in}c@{\hskip 0.1in}c@{\hskip 0.1in}c@{\hskip 0.1in}c@{\hskip 0.1in}c@{\hskip 0.1in}|c@{\hskip 0.1in}}
    \toprule
    ID & OOD & Baseline & Cosine & Gram & MALCOM & Ours & OE \\
    \midrule
    \multirow{8}{*}{\rotatebox[origin=c]{90}{CIFAR-10}}
    & CIFAR-100 & 41.22(1.10) & 60.59(1.20) & 26.90(0.26) & 24.86(0.40) & 47.23(0.47) & 71.55(0.72) \\
    & Food-101 & 47.96(1.93) & 69.06(4.28) & 21.13(0.67) & 20.33(1.19) & 88.50(3.18) & 93.67(0.82) \\
    & iSUN & 59.66(4.76) & 96.08(1.19) & 98.75(0.26) & 94.12(0.57) & 99.71(0.13) & 98.63(0.42) \\
    & LSUNc & 55.13(2.09) & 95.74(0.53) & 88.82(0.72) & 81.39(2.07) & 99.27(0.20) & 98.65(0.40) \\
    & LSUNr & 63.64(3.66) & 96.18(1.52) & 99.32(0.12) & 95.36(0.68) & 99.76(0.15) & 98.85(0.33) \\
    & TINc & 56.01(3.91) & 94.96(0.35) & 96.65(0.47) & 91.40(1.00) & 99.45(0.19) & 96.86(1.12) \\
    & TINr & 54.20(5.27) & 94.17(0.56) & 98.41(0.28) & 94.71(0.61) & 99.44(0.21) & 95.12(1.67) \\
    & SVHN & 35.04(4.77) & 97.76(0.22) & 96.27(0.46) & 94.98(0.80) & 99.37(0.10) & 97.31(1.44) \\
    \midrule
    \multirow{8}{*}{\rotatebox[origin=c]{90}{CIFAR-100}}
    & CIFAR-10 & 20.22(0.74) & 29.96(2.38) & 10.55(0.43) & 1.34(0.08) & 12.88(0.81) & 17.57(0.74) \\
    & Food-101 & 39.42(2.40) & 68.01(4.59) & 14.29(1.31) & 3.70(0.33) & 85.96(3.52) & 49.98(2.37) \\
    & iSUN & 25.85(6.36) & 92.87(0.45) & 96.23(0.63) & 82.96(1.81) & 97.60(0.65) & 59.53(3.48) \\
    & LSUNc & 28.16(1.88) & 84.61(2.56) & 67.16(1.30) & 53.05(1.25) & 96.38(1.02) & 72.87(1.49) \\
    & LSUNr & 29.07(7.10) & 93.39(0.75) & 97.59(0.39) & 85.12(2.06) & 98.65(0.70) & 61.90(3.36) \\
    & TINc & 30.11(5.32) & 93.28(0.84) & 90.65(0.97) & 74.45(1.79) & 97.95(0.92) & 52.03(4.22) \\
    & TINr & 27.06(6.31) & 93.68(0.88) & 95.96(0.39) & 84.26(1.43) & 98.22(0.96) & 46.25(4.88) \\
    & SVHN & 19.85(3.09) & 87.86(3.19) & 77.94(1.95) & 68.30(3.35) & 97.87(0.28) & 54.35(11.76) \\
    \midrule
    \multirow{8}{*}{\rotatebox[origin=c]{90}{SVHN}}
    & CIFAR-10 & 65.90(2.55) & 69.44(8.53) & 81.82(0.58) & 96.68(0.42) & 98.47(0.47) & 99.96(0.03) \\
    & CIFAR-100 & 64.67(2.28) & 67.73(8.77) & 83.96(0.91) & 97.73(0.42) & 97.77(0.42) & 99.89(0.02) \\
    & Food-101 & 64.32(5.90) & 68.66(13.45) & 91.87(1.33) & 99.38(0.08) & 99.18(0.30) & 99.99(0.00) \\
    & iSUN & 70.55(1.66) & 78.05(8.73) & 99.28(0.29) & 99.99(0.01) & 100.00(0.00) & 100.00(0.00) \\
    & LSUNc & 68.18(4.14) & 79.67(7.43) & 93.27(0.88) & 96.20(0.47) & 99.22(0.16) & 99.54(0.07) \\
    & LSUNr & 69.71(1.82) & 74.38(8.83) & 99.47(0.20) & 99.99(0.01) & 100.00(0.00) & 100.00(0.00) \\
    & TINc & 72.40(1.97) & 78.90(10.00) & 98.34(0.54) & 99.67(0.06) & 99.96(0.01) & 99.98(0.00) \\
    & TINr & 72.51(2.40) & 77.33(10.19) & 99.11(0.34) & 99.95(0.03) & 99.99(0.01) & 100.00(0.00) \\
    \midrule
    \multirow{8}{*}{\rotatebox[origin=c]{90}{Food-101}}
    & CIFAR-10 & 13.81(1.58) & 76.34(3.56) & 22.37(0.88) & 36.37(0.64) & 84.03(6.59) & 99.92(0.01) \\
    & CIFAR-100 & 12.30(1.09) & 76.07(3.67) & 31.57(0.54) & 50.41(0.37) & 86.45(5.58) & 99.75(0.04) \\
    & iSUN & 10.67(3.51) & 94.04(2.84) & 93.20(1.16) & 98.33(0.32) & 96.86(1.48) & 39.73(10.71) \\
    & LSUNc & 20.34(2.80) & 93.41(1.59) & 79.20(0.96) & 86.14(0.71) & 74.80(2.40) & 64.09(1.83) \\
    & LSUNr & 11.76(4.43) & 92.80(3.30) & 94.10(1.30) & 98.39(0.36) & 98.22(0.84) & 35.97(10.52) \\
    & TINc & 10.88(2.59) & 91.73(2.40) & 85.96(1.71) & 93.09(0.43) & 85.18(4.05) & 40.75(7.04) \\
    & TINr & 8.24(2.83) & 90.59(2.91) & 90.89(1.31) & 96.12(0.22) & 92.71(2.64) & 33.13(9.00) \\
    & SVHN & 11.56(3.31) & 94.00(2.63) & 92.16(1.44) & 98.30(0.50) & 99.46(0.13) & 94.37(1.95) \\
    \bottomrule
    \end{tabular}
    \end{small}
    \end{center}
\end{table*}

\begin{table*}[tbh]
    \caption{The OOD detection performance in TNR at TPR 95\%. The network architecture is WRN-40-4.}
    \label{table::ood_detection_performance_wrn}
    \begin{center}
    \begin{small}
    \begin{tabular}{@{\hskip 0.1in}c@{\hskip 0.1in}c@{\hskip 0.1in}c@{\hskip 0.1in}c@{\hskip 0.1in}c@{\hskip 0.1in}c@{\hskip 0.1in}c@{\hskip 0.1in}|c@{\hskip 0.1in}}
    \toprule
    ID & OOD & Baseline & Cosine & Gram & MALCOM & Ours & OE \\
    \midrule
    \multirow{8}{*}{\rotatebox[origin=c]{90}{CIFAR-10}}
    & CIFAR-100 & 42.85(1.03) & 56.42(2.64) & 31.97(0.69) & 35.05(0.89) & 48.31(1.64) & 78.48(0.47) \\
    & Food-101 & 45.64(1.20) & 58.92(6.52) & 29.53(1.65) & 37.07(4.97) & 89.94(1.54) & 97.16(0.22) \\
    & iSUN & 55.33(2.18) & 89.82(5.75) & 99.51(0.06) & 97.68(0.35) & 99.83(0.11) & 98.21(0.47) \\
    & LSUNc & 61.08(1.71) & 92.76(2.29) & 92.14(0.57) & 90.26(0.98) & 99.60(0.09) & 99.12(0.20) \\
    & LSUNr & 59.73(1.81) & 91.81(4.17) & 99.62(0.05) & 98.33(0.38) & 99.88(0.10) & 98.46(0.50) \\
    & TINc & 53.83(1.94) & 84.87(8.26) & 97.81(0.15) & 95.44(0.43) & 99.73(0.12) & 96.85(0.62) \\
    & TINr & 48.70(2.22) & 82.53(8.69) & 99.07(0.09) & 96.47(0.35) & 99.65(0.14) & 94.85(1.00) \\
    & SVHN & 63.07(3.73) & 95.12(2.89) & 97.57(0.26) & 94.37(1.44) & 99.45(0.08) & 99.16(0.23) \\
    \midrule
    \multirow{8}{*}{\rotatebox[origin=c]{90}{CIFAR-100}}
    & CIFAR-10 & 22.42(0.71) & 27.65(1.31) & 11.74(0.74) & 9.06(1.16) & 14.37(0.97) & 18.02(0.67) \\
    & Food-101 & 42.92(0.98) & 66.70(1.27) & 19.92(1.07) & 23.30(6.37) & 87.40(2.20) & 47.73(1.95) \\
    & iSUN & 19.97(1.94) & 71.00(6.47) & 96.42(0.11) & 81.17(5.21) & 97.25(0.96) & 39.35(6.75) \\
    & LSUNc & 22.26(1.79) & 72.45(2.48) & 69.24(1.25) & 71.82(0.89) & 97.37(1.05) & 68.10(0.79) \\
    & LSUNr & 22.86(2.24) & 70.31(6.09) & 97.69(0.12) & 81.24(6.84) & 98.44(0.65) & 43.89(6.90) \\
    & TINc & 25.68(1.19) & 77.90(5.80) & 91.20(0.22) & 78.51(4.40) & 98.46(0.76) & 31.98(3.76) \\
    & TINr & 23.41(1.57) & 74.97(6.40) & 95.97(0.39) & 82.81(4.09) & 98.35(0.59) & 27.32(4.71) \\
    & SVHN & 22.02(1.98) & 75.52(9.10) & 82.66(0.53) & 83.88(2.80) & 96.68(0.35) & 66.13(5.96) \\
    \midrule
    \multirow{8}{*}{\rotatebox[origin=c]{90}{SVHN}}
    & CIFAR-10 & 72.46(1.17) & 71.63(10.81) & 84.12(1.70) & 96.23(0.40) & 99.03(0.10) & 99.99(0.00) \\
    & CIFAR-100 & 72.52(1.51) & 69.07(10.59) & 84.71(1.55) & 96.28(0.30) & 98.07(0.29) & 99.98(0.01) \\
    & Food-101 & 76.25(1.57) & 74.39(10.54) & 91.08(1.70) & 98.74(0.05) & 99.57(0.07) & 100.00(0.00) \\
    & iSUN & 72.08(3.97) & 78.10(10.64) & 99.78(0.06) & 99.99(0.00) & 100.00(0.00) & 100.00(0.00) \\
    & LSUNc & 71.37(2.37) & 81.98(6.99) & 94.07(0.31) & 96.56(0.37) & 98.94(0.51) & 99.77(0.04) \\
    & LSUNr & 71.57(3.65) & 74.16(12.45) & 99.87(0.04) & 99.99(0.00) & 100.00(0.00) & 100.00(0.00) \\
    & TINc & 75.71(2.93) & 80.44(10.55) & 98.80(0.17) & 99.65(0.09) & 99.96(0.02) & 100.00(0.00) \\
    & TINr & 74.86(3.12) & 78.12(10.79) & 99.57(0.12) & 99.93(0.03) & 99.97(0.02) & 100.00(0.00) \\
    \midrule
    \multirow{8}{*}{\rotatebox[origin=c]{90}{Food-101}}
    & CIFAR-10 & 13.62(0.24) & 67.86(2.41) & 24.16(0.63) & 37.98(1.27) & 64.90(11.29) & 99.99(0.01) \\
    & CIFAR-100 & 12.74(0.53) & 65.73(1.83) & 33.27(0.76) & 49.83(1.02) & 68.25(10.00) & 100.00(0.00) \\
    & iSUN & 10.90(3.16) & 80.31(8.28) & 94.57(0.97) & 98.15(0.70) & 98.00(0.59) & 78.38(6.39) \\
    & LSUNc & 14.87(2.19) & 88.20(0.98) & 79.27(0.56) & 89.23(0.60) & 73.23(3.02) & 59.19(2.93) \\
    & LSUNr & 11.49(2.99) & 78.11(8.84) & 95.26(0.98) & 98.38(0.62) & 98.40(0.42) & 76.63(8.16) \\
    & TINc & 12.16(2.17) & 81.79(5.08) & 88.52(1.09) & 94.18(0.86) & 88.21(2.04) & 59.22(6.04) \\
    & TINr & 10.57(2.43) & 77.87(7.15) & 92.88(0.92) & 96.19(0.84) & 94.51(1.23) & 67.18(8.29) \\
    & SVHN & 7.80(2.72) & 79.85(5.21) & 96.99(0.80) & 99.12(0.32) & 98.05(0.68) & 99.57(0.17) \\
    \bottomrule
    \end{tabular}
    \end{small}
    \end{center}
\end{table*}

\begin{table*}[tbh]
    \caption{The OOD detection performance in AUROC. The network architecture is DenseNet-100-12.}
    \label{table::ood_detection_performance_dense_auroc}
    \begin{center}
    \begin{small}
    \begin{tabular}{@{\hskip 0.1in}c@{\hskip 0.1in}c@{\hskip 0.1in}c@{\hskip 0.1in}c@{\hskip 0.1in}c@{\hskip 0.1in}c@{\hskip 0.1in}c@{\hskip 0.1in}|c@{\hskip 0.1in}}
    \toprule
    ID & OOD & Baseline & Cosine & Gram & MALCOM & Ours & OE \\
    \midrule
    \multirow{8}{*}{\rotatebox[origin=c]{90}{CIFAR-10}}
    & CIFAR-100 & 89.48(0.15) & 90.21(0.22) & 72.63(0.43) & 71.15(0.30) & 86.06(0.42) & 94.19(0.13) \\
    & Food-101 & 91.75(0.48) & 93.74(1.06) & 72.79(0.54) & 81.03(1.08) & 97.68(0.65) & 98.57(0.09) \\
    & iSUN & 94.44(0.78) & 99.10(0.25) & 99.73(0.05) & 98.78(0.10) & 99.88(0.03) & 99.45(0.13) \\
    & LSUNc & 93.52(0.50) & 99.09(0.12) & 97.46(0.09) & 96.46(0.41) & 99.82(0.04) & 99.56(0.07) \\
    & LSUNr & 95.10(0.53) & 99.09(0.31) & 99.84(0.03) & 98.92(0.12) & 99.89(0.04) & 99.50(0.11) \\
    & TINc & 93.66(0.87) & 98.87(0.08) & 99.23(0.08) & 98.30(0.17) & 99.85(0.05) & 99.10(0.23) \\
    & TINr & 93.30(1.06) & 98.72(0.12) & 99.63(0.06) & 98.88(0.11) & 99.84(0.05) & 98.79(0.32) \\
    & SVHN & 85.36(4.85) & 99.47(0.07) & 99.15(0.09) & 98.86(0.16) & 99.83(0.03) & 99.22(0.30) \\
    \midrule
    \multirow{8}{*}{\rotatebox[origin=c]{90}{CIFAR-100}}
    & CIFAR-10 & 77.86(0.28) & 78.68(1.04) & 63.93(0.47) & 45.32(0.32) & 74.46(0.46) & 78.98(0.45) \\
    & Food-101 & 86.97(0.69) & 94.76(0.64) & 70.86(0.95) & 71.24(1.52) & 97.27(0.65) & 91.36(0.53) \\
    & iSUN & 80.25(1.61) & 98.48(0.09) & 99.14(0.10) & 97.04(0.32) & 99.47(0.15) & 90.21(1.01) \\
    & LSUNc & 80.44(0.84) & 96.89(0.56) & 92.43(0.25) & 91.54(0.49) & 99.23(0.18) & 94.02(0.40) \\
    & LSUNr & 81.99(1.55) & 98.64(0.15) & 99.45(0.06) & 97.25(0.36) & 99.69(0.13) & 90.98(1.19) \\
    & TINc & 81.94(1.53) & 98.60(0.20) & 97.97(0.14) & 95.61(0.37) & 99.55(0.17) & 88.04(1.34) \\
    & TINr & 80.87(2.17) & 98.66(0.20) & 99.10(0.05) & 97.18(0.30) & 99.62(0.18) & 86.17(1.64) \\
    & SVHN & 77.28(2.12) & 97.70(0.49) & 95.39(0.38) & 94.68(0.39) & 99.46(0.05) & 91.52(2.30) \\
    \midrule
    \multirow{8}{*}{\rotatebox[origin=c]{90}{SVHN}}
    & CIFAR-10 & 91.55(1.35) & 93.14(1.93) & 95.61(0.18) & 98.99(0.06) & 99.57(0.10) & 99.99(0.00) \\
    & CIFAR-100 & 90.95(1.18) & 92.60(2.16) & 96.38(0.19) & 99.17(0.06) & 99.39(0.09) & 99.97(0.00) \\
    & Food-101 & 90.33(3.62) & 92.58(3.88) & 98.03(0.35) & 99.63(0.02) & 99.71(0.06) & 99.99(0.00) \\
    & iSUN & 92.88(0.93) & 95.04(2.19) & 99.80(0.08) & 99.96(0.01) & 99.97(0.02) & 100.00(0.00) \\
    & LSUNc & 91.23(2.38) & 94.81(2.33) & 98.57(0.15) & 99.07(0.09) & 99.77(0.05) & 99.88(0.02) \\
    & LSUNr & 92.70(1.16) & 94.14(2.33) & 99.84(0.05) & 99.96(0.01) & 99.97(0.02) & 100.00(0.00) \\
    & TINc & 93.32(1.13) & 95.32(2.52) & 99.55(0.11) & 99.85(0.02) & 99.96(0.02) & 99.99(0.00) \\
    & TINr & 93.43(1.35) & 94.86(2.59) & 99.76(0.08) & 99.93(0.01) & 99.97(0.02) & 100.00(0.00) \\
    \midrule
    \multirow{8}{*}{\rotatebox[origin=c]{90}{Food-101}}
    & CIFAR-10 & 67.71(1.13) & 95.57(0.62) & 71.75(0.70) & 83.55(0.54) & 95.56(1.95) & 99.98(0.00) \\
    & CIFAR-100 & 65.13(0.82) & 95.46(0.67) & 76.99(0.23) & 87.41(0.33) & 96.07(1.69) & 99.92(0.02) \\
    & iSUN & 62.06(4.88) & 98.83(0.51) & 98.53(0.25) & 99.49(0.08) & 99.18(0.34) & 79.95(7.14) \\
    & LSUNc & 70.21(1.73) & 98.73(0.29) & 94.36(0.28) & 96.92(0.14) & 92.10(0.91) & 88.06(0.67) \\
    & LSUNr & 63.85(4.67) & 98.62(0.58) & 98.60(0.26) & 99.47(0.09) & 99.54(0.19) & 79.00(6.89) \\
    & TINc & 61.37(3.28) & 98.38(0.49) & 96.87(0.38) & 98.48(0.10) & 95.72(1.04) & 78.89(4.43) \\
    & TINr & 58.90(4.24) & 98.18(0.55) & 97.87(0.30) & 99.09(0.07) & 97.92(0.62) & 75.42(6.11) \\
    & SVHN & 65.62(4.69) & 98.82(0.44) & 98.33(0.21) & 99.30(0.10) & 99.86(0.04) & 98.20(0.58) \\
    \bottomrule
    \end{tabular}
    \end{small}
    \end{center}
\end{table*}

\begin{table*}[tbh]
    \caption{The OOD detection performance in AUROC. The network architecture is WRN-40-4.}
    \label{table::ood_detection_performance_wrn_auroc}
    \begin{center}
    \begin{small}
    \begin{tabular}{@{\hskip 0.1in}c@{\hskip 0.1in}c@{\hskip 0.1in}c@{\hskip 0.1in}c@{\hskip 0.1in}c@{\hskip 0.1in}c@{\hskip 0.1in}c@{\hskip 0.1in}|c@{\hskip 0.1in}}
    \toprule
    ID & OOD & Baseline & Cosine & Gram & MALCOM & Ours & OE \\
    \midrule
    \multirow{8}{*}{\rotatebox[origin=c]{90}{CIFAR-10}}
    & CIFAR-100 & 86.54(0.13) & 88.87(0.81) & 79.50(0.24) & 83.03(0.68) & 85.56(0.29) & 95.65(0.07) \\
    & Food-101 & 88.16(0.40) & 91.54(1.71) & 81.71(0.66) & 88.15(1.29) & 98.00(0.27) & 98.85(0.05) \\
    & iSUN & 91.99(0.60) & 98.02(1.00) & 99.85(0.01) & 99.35(0.08) & 99.84(0.06) & 98.75(0.12) \\
    & LSUNc & 93.43(0.41) & 98.57(0.43) & 98.39(0.09) & 98.17(0.18) & 99.83(0.05) & 99.11(0.04) \\
    & LSUNr & 93.26(0.53) & 98.39(0.72) & 99.89(0.01) & 99.41(0.09) & 99.85(0.07) & 98.77(0.12) \\
    & TINc & 90.98(0.51) & 97.06(1.60) & 99.50(0.03) & 98.97(0.09) & 99.84(0.05) & 98.64(0.10) \\
    & TINr & 89.22(1.06) & 96.47(1.76) & 99.77(0.02) & 99.18(0.06) & 99.80(0.07) & 98.21(0.15) \\
    & SVHN & 94.09(0.73) & 98.94(0.72) & 99.47(0.06) & 98.88(0.27) & 99.81(0.03) & 98.91(0.26) \\
    \midrule
    \multirow{8}{*}{\rotatebox[origin=c]{90}{CIFAR-100}}
    & CIFAR-10 & 78.82(0.43) & 77.84(0.85) & 67.34(0.21) & 70.57(1.11) & 72.16(1.33) & 81.89(0.12) \\
    & Food-101 & 88.24(0.37) & 94.41(0.15) & 78.05(0.46) & 86.33(1.95) & 97.50(0.38) & 92.46(0.19) \\
    & iSUN & 74.04(1.22) & 94.52(1.29) & 99.10(0.03) & 96.40(0.88) & 99.41(0.15) & 89.53(1.87) \\
    & LSUNc & 78.34(0.80) & 94.46(0.68) & 93.39(0.20) & 94.97(0.25) & 99.41(0.18) & 94.59(0.18) \\
    & LSUNr & 75.45(1.03) & 94.48(1.27) & 99.39(0.02) & 96.44(1.12) & 99.63(0.11) & 90.83(1.56) \\
    & TINc & 76.43(1.11) & 95.92(1.07) & 98.09(0.08) & 96.11(0.68) & 99.64(0.13) & 86.90(1.64) \\
    & TINr & 74.20(1.23) & 95.26(1.31) & 99.05(0.07) & 96.66(0.71) & 99.60(0.10) & 85.34(1.95) \\
    & SVHN & 78.87(1.30) & 95.62(1.58) & 96.34(0.17) & 96.94(0.53) & 99.21(0.08) & 94.52(0.76) \\
    \midrule
    \multirow{8}{*}{\rotatebox[origin=c]{90}{SVHN}}
    & CIFAR-10 & 92.56(0.48) & 93.62(2.62) & 96.34(0.41) & 99.01(0.09) & 99.67(0.05) & 99.99(0.00) \\
    & CIFAR-100 & 92.49(0.59) & 92.61(2.68) & 96.59(0.37) & 99.04(0.07) & 99.45(0.06) & 99.99(0.00) \\
    & Food-101 & 94.04(0.62) & 94.40(2.14) & 97.88(0.37) & 99.56(0.04) & 99.82(0.02) & 99.99(0.00) \\
    & iSUN & 92.92(1.43) & 95.14(2.69) & 99.92(0.02) & 99.98(0.01) & 99.92(0.03) & 99.99(0.00) \\
    & LSUNc & 92.12(1.00) & 95.65(1.76) & 98.73(0.02) & 99.20(0.07) & 99.69(0.10) & 99.93(0.00) \\
    & LSUNr & 92.44(1.51) & 94.08(3.47) & 99.94(0.01) & 99.99(0.00) & 99.90(0.03) & 99.99(0.00) \\
    & TINc & 94.09(1.15) & 95.76(2.44) & 99.66(0.05) & 99.88(0.02) & 99.92(0.02) & 99.99(0.00) \\
    & TINr & 93.83(1.08) & 95.22(2.42) & 99.85(0.03) & 99.95(0.01) & 99.90(0.03) & 99.99(0.00) \\
    \midrule
    \multirow{8}{*}{\rotatebox[origin=c]{90}{Food-101}}
    & CIFAR-10 & 68.62(0.13) & 94.06(0.50) & 74.00(0.44) & 83.32(0.64) & 90.23(3.54) & 100.00(0.00) \\
    & CIFAR-100 & 67.00(0.32) & 93.40(0.47) & 79.18(0.51) & 86.75(0.52) & 90.90(3.12) & 100.00(0.00) \\
    & iSUN & 66.28(2.72) & 96.53(1.45) & 98.81(0.22) & 99.53(0.15) & 99.49(0.14) & 92.90(2.27) \\
    & LSUNc & 68.23(2.91) & 97.78(0.20) & 94.44(0.29) & 97.63(0.15) & 91.92(0.93) & 85.89(0.98) \\
    & LSUNr & 66.29(2.41) & 96.13(1.61) & 98.90(0.20) & 99.52(0.16) & 99.61(0.11) & 92.35(2.73) \\
    & TINc & 66.55(1.49) & 96.64(0.86) & 97.44(0.27) & 98.72(0.16) & 96.68(0.63) & 85.36(2.59) \\
    & TINr & 65.36(1.84) & 95.92(1.31) & 98.32(0.23) & 99.10(0.19) & 98.51(0.37) & 88.41(3.10) \\
    & SVHN & 63.36(2.73) & 96.25(1.15) & 99.23(0.17) & 99.68(0.10) & 99.49(0.17) & 99.87(0.07) \\
    \bottomrule
    \end{tabular}
    \end{small}
    \end{center}
\end{table*}

\begin{table*}[tbh]
    \caption{The OOD detection performance in AUPR. The network architecture is DenseNet-100-12.}
    \label{table::ood_detection_performance_dense_aupr}
    \begin{center}
    \begin{small}
    \begin{tabular}{@{\hskip 0.1in}c@{\hskip 0.1in}c@{\hskip 0.1in}c@{\hskip 0.1in}c@{\hskip 0.1in}c@{\hskip 0.1in}c@{\hskip 0.1in}c@{\hskip 0.1in}|c@{\hskip 0.1in}}
    \toprule
    ID & OOD & Baseline & Cosine & Gram & MALCOM & Ours & OE \\
    \midrule
    \multirow{8}{*}{\rotatebox[origin=c]{90}{CIFAR-10}}
    & CIFAR-100 & 90.64(0.11) & 88.44(0.44) & 62.57(0.68) & 69.23(0.37) & 84.77(0.50) & 93.96(0.18) \\
    & Food-101 & 87.50(0.80) & 87.36(2.25) & 44.41(1.04) & 71.27(1.66) & 95.24(1.35) & 97.37(0.21) \\
    & iSUN & 95.95(0.55) & 99.15(0.24) & 99.72(0.05) & 99.01(0.08) & 99.90(0.02) & 99.58(0.10) \\
    & LSUNc & 94.79(0.44) & 98.99(0.14) & 96.21(0.20) & 96.74(0.39) & 99.81(0.04) & 99.60(0.07) \\
    & LSUNr & 96.10(0.41) & 99.05(0.32) & 99.81(0.02) & 99.05(0.10) & 99.90(0.03) & 99.58(0.09) \\
    & TINc & 94.92(0.75) & 98.81(0.08) & 98.85(0.10) & 98.42(0.17) & 99.85(0.05) & 99.21(0.19) \\
    & TINr & 94.61(0.85) & 98.62(0.10) & 99.46(0.14) & 98.94(0.10) & 99.84(0.05) & 98.91(0.29) \\
    & SVHN & 70.78(14.06) & 98.65(0.11) & 96.86(0.20) & 97.56(0.28) & 99.55(0.06) & 98.60(0.43) \\
    \midrule
    \multirow{8}{*}{\rotatebox[origin=c]{90}{CIFAR-100}}
    & CIFAR-10 & 80.41(0.29) & 74.76(1.33) & 60.99(0.50) & 49.47(0.25) & 76.94(0.44) & 81.96(0.31) \\
    & Food-101 & 80.20(0.83) & 91.52(0.90) & 50.39(1.22) & 61.04(2.36) & 94.65(1.22) & 86.62(0.77) \\
    & iSUN & 83.84(1.22) & 98.57(0.07) & 99.05(0.11) & 97.68(0.24) & 99.49(0.15) & 90.87(0.93) \\
    & LSUNc & 82.32(0.75) & 96.66(0.64) & 91.15(0.29) & 92.68(0.53) & 99.21(0.18) & 93.82(0.47) \\
    & LSUNr & 84.11(1.18) & 98.66(0.14) & 99.38(0.08) & 97.68(0.30) & 99.69(0.15) & 90.87(1.34) \\
    & TINc & 83.86(1.55) & 98.57(0.18) & 97.54(0.17) & 96.12(0.35) & 99.54(0.17) & 87.86(1.16) \\
    & TINr & 83.08(1.97) & 98.62(0.16) & 98.92(0.05) & 97.46(0.28) & 99.60(0.20) & 86.05(1.28) \\
    & SVHN & 66.54(3.26) & 95.13(0.82) & 88.85(1.00) & 91.21(0.54) & 98.45(0.15) & 85.74(3.22) \\
    \midrule
    \multirow{8}{*}{\rotatebox[origin=c]{90}{SVHN}}
    & CIFAR-10 & 95.28(1.25) & 96.84(0.88) & 89.36(0.49) & 99.66(0.02) & 99.82(0.05) & 99.99(0.00) \\
    & CIFAR-100 & 94.69(1.19) & 96.48(1.06) & 91.46(0.41) & 99.72(0.02) & 99.71(0.06) & 99.99(0.00) \\
    & Food-101 & 87.14(6.69) & 92.51(3.88) & 97.62(0.46) & 99.72(0.02) & 99.74(0.06) & 100.00(0.00) \\
    & iSUN & 96.43(0.74) & 97.90(1.00) & 99.31(0.26) & 99.99(0.00) & 99.99(0.00) & 100.00(0.00) \\
    & LSUNc & 94.40(2.42) & 97.30(1.37) & 97.09(0.31) & 99.67(0.03) & 99.90(0.02) & 99.95(0.01) \\
    & LSUNr & 95.95(0.99) & 97.22(1.21) & 99.48(0.18) & 99.99(0.00) & 99.99(0.01) & 100.00(0.00) \\
    & TINc & 96.20(1.11) & 97.82(1.24) & 98.85(0.30) & 99.95(0.00) & 99.99(0.01) & 100.00(0.00) \\
    & TINr & 96.29(1.24) & 97.59(1.27) & 99.30(0.24) & 99.98(0.00) & 99.99(0.01) & 100.00(0.00) \\
    \midrule
    \multirow{8}{*}{\rotatebox[origin=c]{90}{Food-101}}
    & CIFAR-10 & 84.13(0.53) & 98.03(0.27) & 83.10(0.51) & 92.46(0.31) & 97.67(1.03) & 99.99(0.00) \\
    & CIFAR-100 & 82.47(0.35) & 97.97(0.28) & 86.38(0.27) & 94.18(0.19) & 97.89(0.90) & 99.95(0.01) \\
    & iSUN & 82.43(2.65) & 99.53(0.20) & 99.37(0.11) & 99.83(0.02) & 99.62(0.15) & 91.30(3.31) \\
    & LSUNc & 84.66(0.86) & 99.41(0.13) & 96.58(0.21) & 98.56(0.08) & 95.60(0.53) & 93.47(0.34) \\
    & LSUNr & 82.01(2.50) & 99.39(0.25) & 99.33(0.12) & 99.80(0.03) & 99.77(0.09) & 90.17(3.35) \\
    & TINc & 80.06(1.82) & 99.26(0.23) & 98.36(0.21) & 99.35(0.05) & 97.67(0.53) & 89.33(2.18) \\
    & TINr & 78.94(2.38) & 99.17(0.24) & 98.89(0.17) & 99.62(0.03) & 98.86(0.29) & 87.98(2.97) \\
    & SVHN & 67.02(4.30) & 98.74(0.43) & 98.22(0.23) & 99.41(0.09) & 99.82(0.05) & 97.34(0.80) \\
    \bottomrule
    \end{tabular}
    \end{small}
    \end{center}
\end{table*}

\begin{table*}[tbh]
    \caption{The OOD detection performance in AUPR. The network architecture is WRN-40-4.}
    \label{table::ood_detection_performance_wrn_aupr}
    \begin{center}
    \begin{small}
    \begin{tabular}{@{\hskip 0.1in}c@{\hskip 0.1in}c@{\hskip 0.1in}c@{\hskip 0.1in}c@{\hskip 0.1in}c@{\hskip 0.1in}c@{\hskip 0.1in}c@{\hskip 0.1in}|c@{\hskip 0.1in}}
    \toprule
    ID & OOD & Baseline & Cosine & Gram & MALCOM & Ours & OE \\
    \midrule
    \multirow{8}{*}{\rotatebox[origin=c]{90}{CIFAR-10}}
    & CIFAR-100 & 83.29(0.16) & 86.81(0.94) & 74.26(0.45) & 82.88(0.80) & 83.03(0.39) & 95.81(0.08) \\
    & Food-101 & 72.55(1.34) & 84.01(2.99) & 64.32(1.61) & 80.89(2.02) & 96.38(0.65) & 98.39(0.05) \\
    & iSUN & 92.56(0.79) & 98.20(0.89) & 99.85(0.01) & 99.48(0.06) & 99.88(0.05) & 99.20(0.07) \\
    & LSUNc & 93.19(0.52) & 98.43(0.46) & 98.19(0.11) & 98.38(0.15) & 99.84(0.05) & 99.37(0.03) \\
    & LSUNr & 93.26(0.72) & 98.40(0.68) & 99.88(0.01) & 99.50(0.07) & 99.87(0.06) & 99.18(0.09) \\
    & TINc & 90.08(0.85) & 96.98(1.62) & 99.44(0.04) & 99.08(0.07) & 99.86(0.05) & 99.00(0.08) \\
    & TINr & 87.95(1.76) & 96.31(1.76) & 99.71(0.05) & 99.24(0.05) & 99.82(0.07) & 98.54(0.12) \\
    & SVHN & 87.76(2.46) & 97.03(2.03) & 98.45(0.16) & 97.44(0.57) & 99.54(0.07) & 98.69(0.20) \\
    \midrule
    \multirow{8}{*}{\rotatebox[origin=c]{90}{CIFAR-100}}
    & CIFAR-10 & 80.00(0.75) & 74.53(1.46) & 65.28(0.29) & 71.71(1.11) & 71.17(1.74) & 85.51(0.07) \\
    & Food-101 & 80.92(0.69) & 90.60(0.29) & 63.42(0.81) & 81.47(2.34) & 95.11(0.73) & 89.94(0.17) \\
    & iSUN & 75.65(1.40) & 95.04(1.06) & 99.00(0.04) & 97.05(0.68) & 99.48(0.14) & 92.24(1.47) \\
    & LSUNc & 79.43(1.01) & 94.23(0.81) & 92.49(0.19) & 95.54(0.29) & 99.40(0.17) & 95.47(0.17) \\
    & LSUNr & 75.08(0.79) & 94.64(1.17) & 99.36(0.02) & 96.82(0.96) & 99.64(0.11) & 92.73(1.28) \\
    & TINc & 75.13(2.66) & 95.95(0.98) & 97.81(0.08) & 96.53(0.56) & 99.64(0.13) & 89.06(1.62) \\
    & TINr & 72.70(2.85) & 95.26(1.22) & 98.92(0.07) & 96.87(0.62) & 99.60(0.10) & 87.62(1.79) \\
    & SVHN & 64.30(4.74) & 91.52(2.61) & 90.93(0.48) & 94.01(0.98) & 97.97(0.20) & 91.60(0.93) \\
    \midrule
    \multirow{8}{*}{\rotatebox[origin=c]{90}{SVHN}}
    & CIFAR-10 & 95.39(0.44) & 97.10(1.20) & 91.23(0.85) & 99.66(0.03) & 99.87(0.02) & 100.00(0.00) \\
    & CIFAR-100 & 95.34(0.57) & 96.48(1.27) & 92.04(0.80) & 99.67(0.02) & 99.77(0.02) & 100.00(0.00) \\
    & Food-101 & 92.42(1.20) & 94.46(1.79) & 97.56(0.43) & 99.64(0.03) & 99.84(0.02) & 99.99(0.00) \\
    & iSUN & 96.38(1.00) & 98.02(1.16) & 99.75(0.06) & 99.99(0.00) & 99.97(0.01) & 100.00(0.00) \\
    & LSUNc & 95.21(0.93) & 97.89(0.86) & 97.41(0.07) & 99.72(0.02) & 99.87(0.04) & 99.96(0.00) \\
    & LSUNr & 95.50(1.22) & 97.26(1.70) & 99.83(0.03) & 99.99(0.00) & 99.97(0.01) & 100.00(0.00) \\
    & TINc & 96.75(0.95) & 98.11(1.10) & 99.21(0.12) & 99.96(0.01) & 99.97(0.01) & 100.00(0.00) \\
    & TINr & 96.52(0.90) & 97.85(1.08) & 99.63(0.07) & 99.98(0.01) & 99.97(0.01) & 100.00(0.00) \\
    \midrule
    \multirow{8}{*}{\rotatebox[origin=c]{90}{Food-101}}
    & CIFAR-10 & 84.66(0.13) & 97.36(0.23) & 84.80(0.37) & 92.28(0.38) & 95.03(1.77) & 100.00(0.00) \\
    & CIFAR-100 & 83.56(0.20) & 96.97(0.26) & 88.02(0.44) & 93.76(0.31) & 95.29(1.62) & 100.00(0.00) \\
    & iSUN & 85.07(1.36) & 98.65(0.56) & 99.47(0.10) & 99.84(0.05) & 99.77(0.07) & 96.50(1.09) \\
    & LSUNc & 84.01(1.81) & 98.98(0.09) & 96.68(0.23) & 98.93(0.07) & 95.52(0.50) & 92.33(0.50) \\
    & LSUNr & 83.73(1.28) & 98.32(0.70) & 99.45(0.10) & 99.81(0.06) & 99.81(0.06) & 95.85(1.40) \\
    & TINc & 83.39(0.83) & 98.49(0.38) & 98.64(0.16) & 99.45(0.07) & 98.19(0.36) & 91.96(1.37) \\
    & TINr & 82.98(0.92) & 98.18(0.57) & 99.10(0.12) & 99.61(0.09) & 99.21(0.20) & 93.58(1.60) \\
    & SVHN & 65.59(2.17) & 96.06(1.28) & 99.19(0.17) & 99.69(0.09) & 99.35(0.23) & 99.80(0.11) \\
    \bottomrule
    \end{tabular}
    \end{small}
    \end{center}
\end{table*}

\end{document}